\documentclass[runningheads]{llncs}
\usepackage{graphicx}
\usepackage{tikz}
\usepackage{comment}
\usepackage{amsmath,amssymb} % define this before the line numbering.
\usepackage{color}

\usepackage{subfig, mathtools, caption, float}
\usepackage[pagebackref=true,breaklinks=true,letterpaper=true,colorlinks,bookmarks=false]{hyperref}

\def\best{\bf \cellcolor[gray]{0.85}}
\def\secbest{\cellcolor[gray]{0.92} }

\usepackage{color}
\usepackage{colortbl}
\newcommand{\bgGray}[1]{\cellcolor[gray]{0.85} #1}

\usepackage{array}
\newcolumntype{C}[1]{>{\centering\arraybackslash}m{#1}}
\newcolumntype{R}[1]{>{\raggedleft\arraybackslash}m{#1}}
\newcolumntype{P}[1]{>{\raggedright\arraybackslash}p{#1}}
\newcolumntype{M}[1]{>{\centering\arraybackslash}m{#1}}

\newcommand{\ie}{\textit{i}.\textit{e}.}
\newcommand{\eg}{\textit{e}.\textit{g}.}

\begin{document}
% \linenumbers
\pagestyle{headings}
\mainmatter
\def\ECCVSubNumber{5211}  % Insert your submission number here

\title{EfficientFCN: Holistically-guided Decoding for Semantic Segmentation} % Replace with your title

\begin{comment}
\titlerunning{ECCV-20 submission ID \ECCVSubNumber} 
\authorrunning{ECCV-20 submission ID \ECCVSubNumber} 
\author{Anonymous ECCV submission}
\institute{Paper ID \ECCVSubNumber}
\end{comment}
%******************

%\begin{comment}
\titlerunning{Holistically-guided Decoding for Semantic Segmentation}
\author{Jianbo Liu\inst{1} \and
Junjun He\inst{2} \and
Jiawei Zhang\inst{3} \and
Jimmy S. Ren\inst{3} \and
Hongsheng Li\inst{1}}
\authorrunning{J. Liu et al. }
\institute{CUHK-SenseTime Joint Laboratory, The Chinese University of Hong Kong \and
Shenzhen Key Lab of Computer Vision and Pattern Recognition, Shenzhen Institutes of Advanced Technology,
Chinese Academy of Sciences\and
SenseTime Research \\
\email{\{liujianbo@link, hsli@ee\}.cuhk.edu.hk}}
%\end{comment}
%******************
\maketitle

\begin{abstract}
Both performance and efficiency are important to semantic segmentation. State-of-the-art 
semantic segmentation algorithms are mostly based on dilated Fully Convolutional Networks 
(dilatedFCN), which adopt dilated convolutions in the backbone networks to extract 
high-resolution feature maps for achieving high-performance segmentation performance. However,
due to many convolution operations are conducted on the high-resolution feature maps, 
such dilatedFCN-based methods result in large computational complexity and memory consumption. 
To balance the performance and efficiency, there also exist encoder-decoder structures that 
gradually recover the spatial information by combining multi-level feature maps from the encoder.
However, the performances of existing encoder-decoder methods are far from comparable with the dilatedFCN-based methods.
In this paper, we propose the EfficientFCN, whose backbone is a common ImageNet pretrained 
network without any dilated convolution. A holistically-guided decoder is introduced to obtain 
the high-resolution semantic-rich feature maps via the multi-scale features from the encoder.
The decoding task is converted to novel codebook generation and codeword assembly task, which 
takes advantages of the high-level and low-level features from the encoder. Such a framework 
achieves comparable or even better performance than state-of-the-art methods with only 1/3 of the computational cost. Extensive experiments on PASCAL Context, PASCAL VOC, ADE20K validate the effectiveness of the proposed EfficientFCN.
% any  an effective holistically-guided decoding method for semantic segmentation to get the competitive accuracy  result without the dilated convolution in the backbone network and achieve more than XXX times faster efficiency.
%\dots
\end{abstract}

\section{Introduction}

\begin{figure}[!t]
\centering
\begin{center}
\begin{tabular}{C{5.0cm}C{5.0cm}}
    \includegraphics[height=3.8cm]{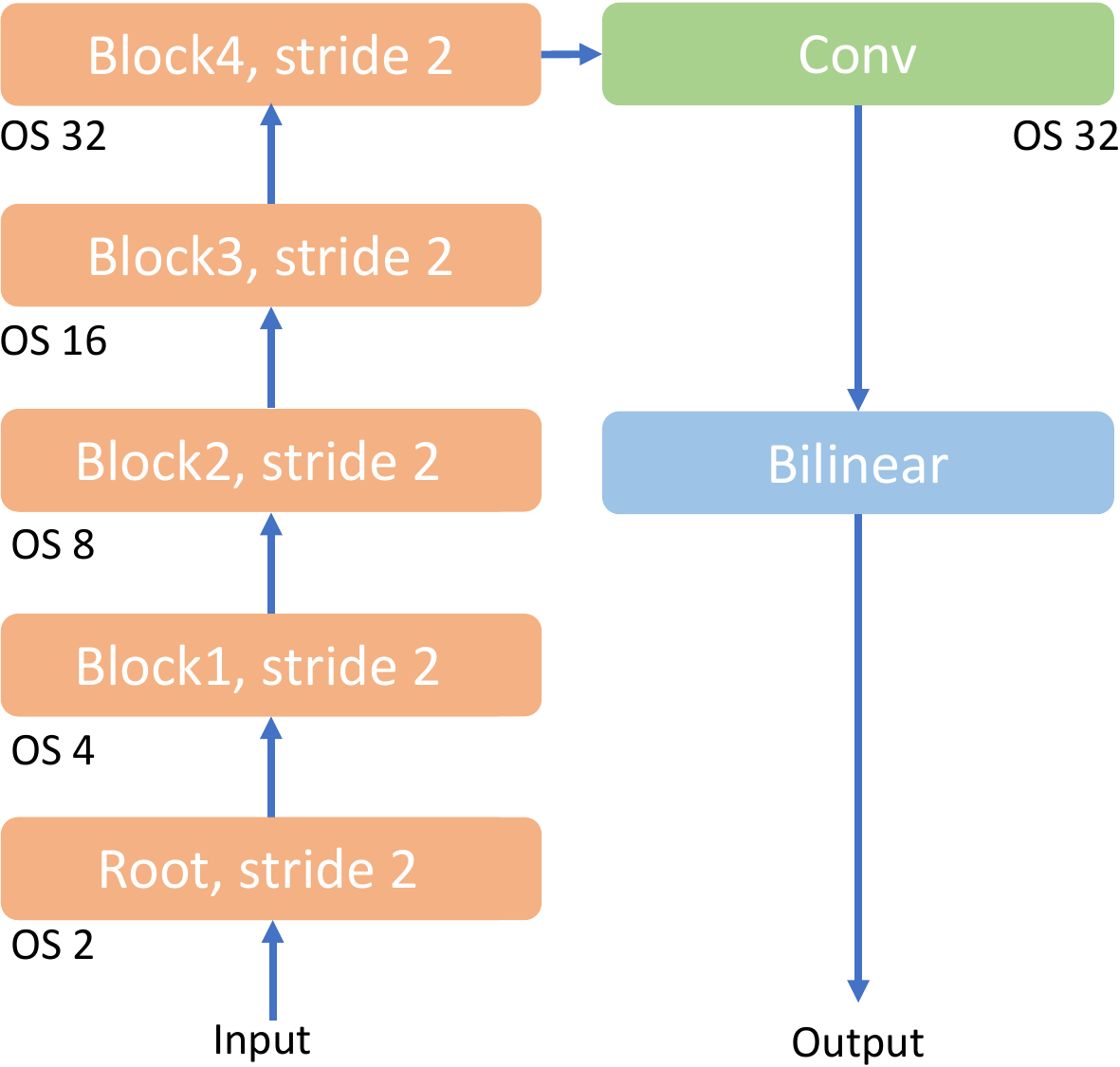} &
    \includegraphics[height=3.8cm]{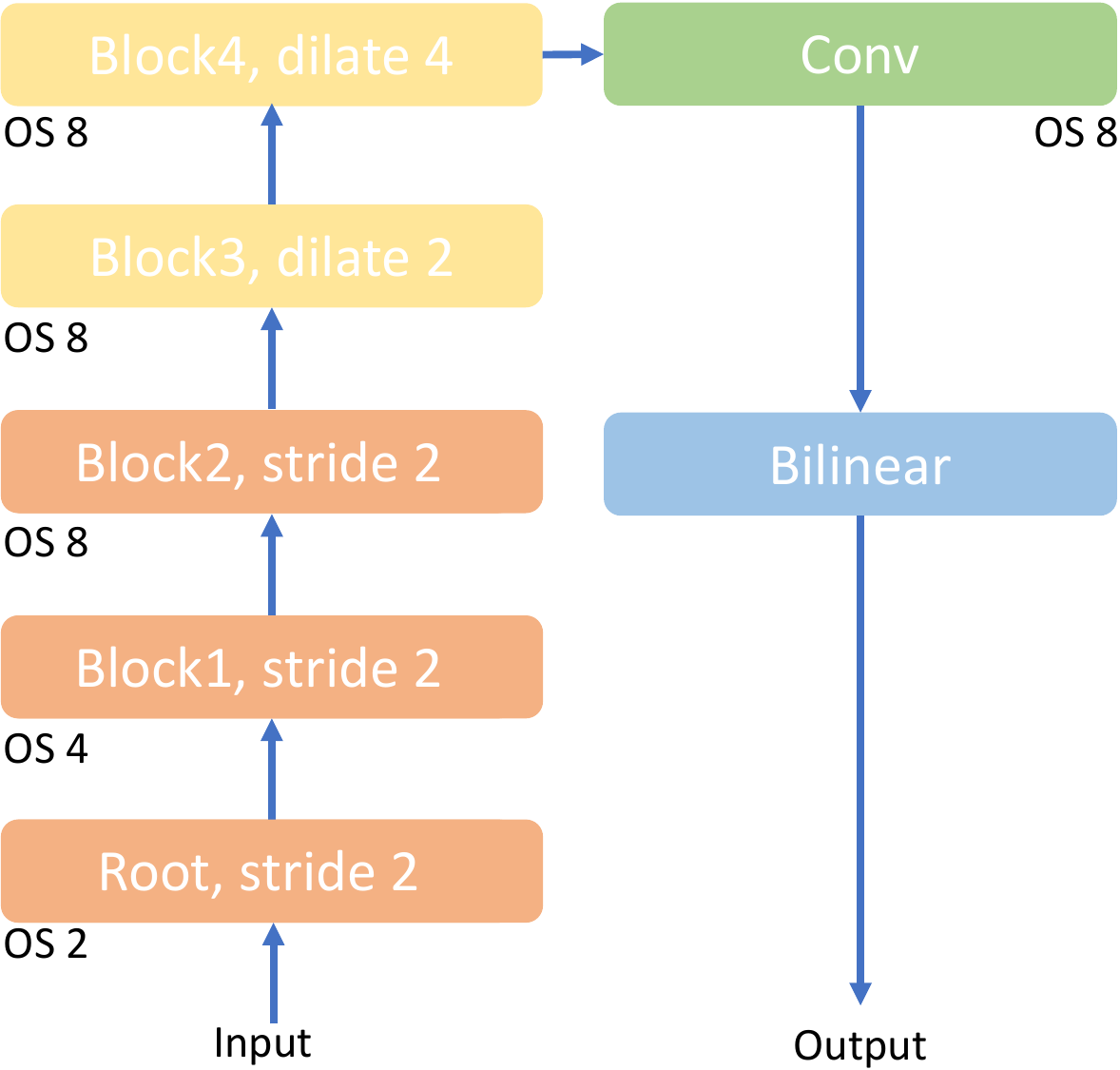} \\
    \centering (a) FCN & (b) DilatedFCN \\ \\
    \includegraphics[height=3.8cm]{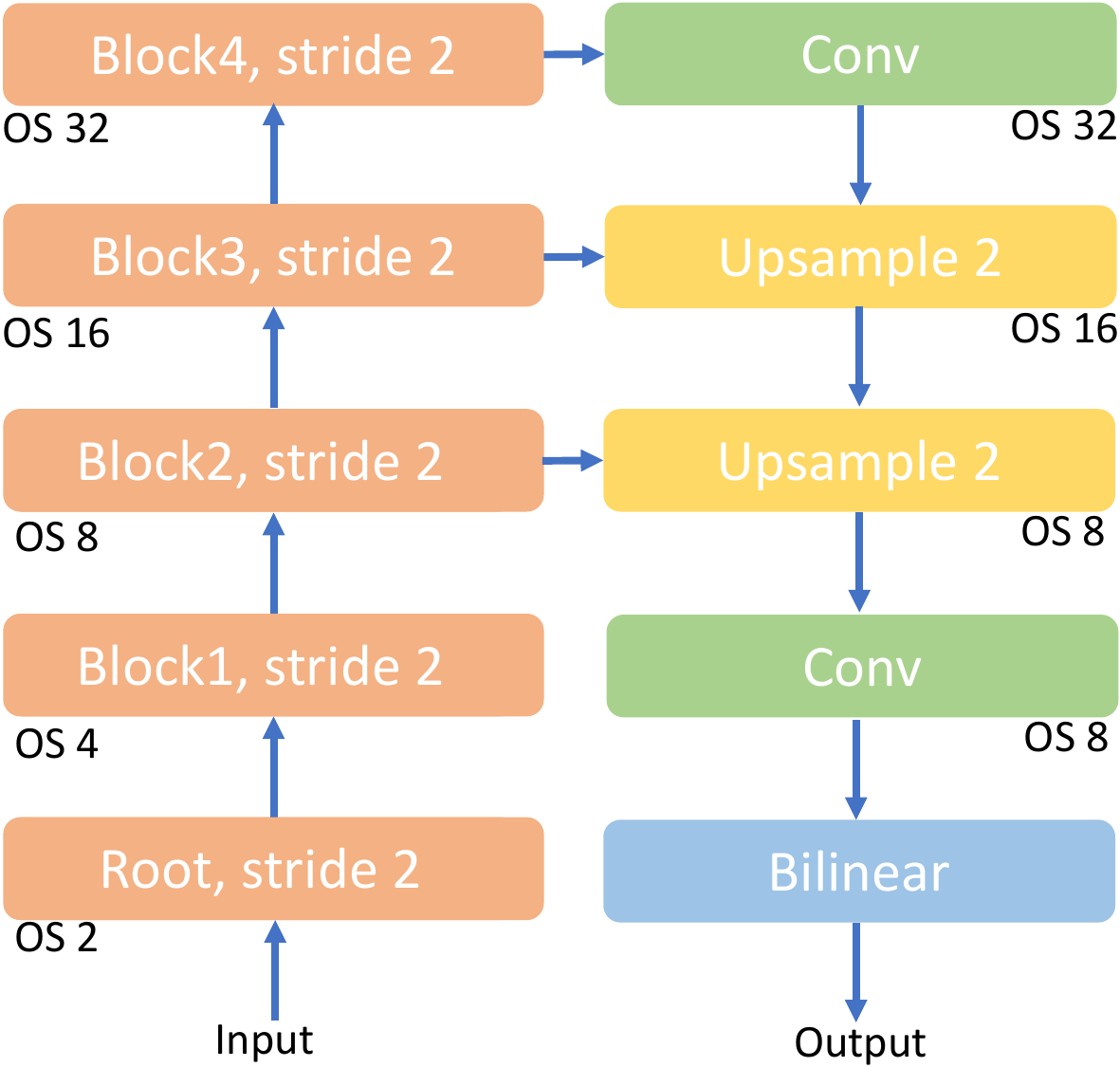} &
    \includegraphics[height=3.8cm]{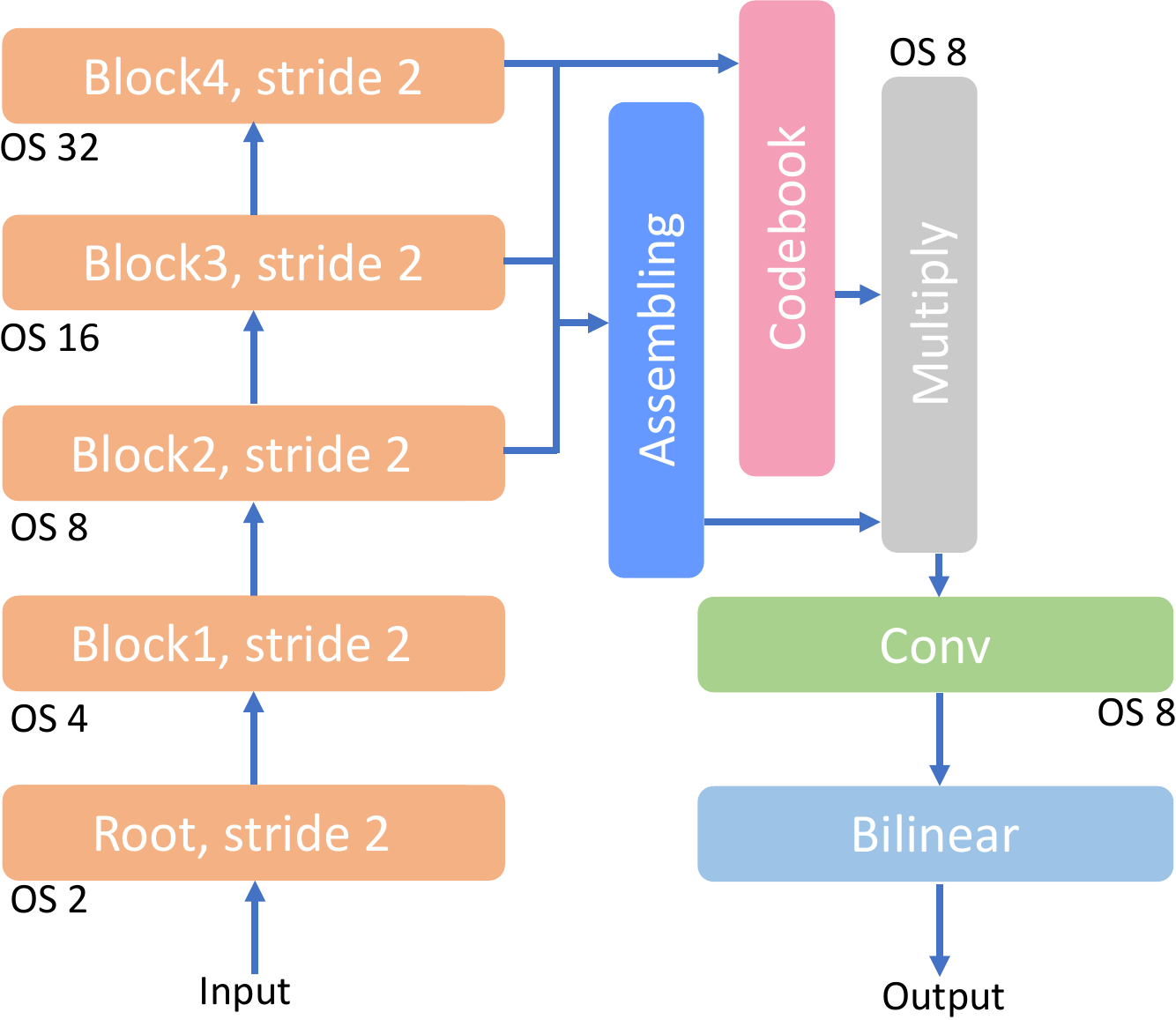} \\
    \centering (c) Encoder-Decoder & (d) EfficientFCN 
\end{tabular}
\end{center}
\caption{Different architectures for semantic segmentation. (a) the original FCN with output stride (OS)=32. (b). DilatedFCN based methods sacrifice efficiency and exploit the dilated convolution with stride 2 and 4 in the last two stages to generate high-resolution feature maps. (c)Encoder-Decoder methods employ the U-Net structure to recover the high-resolution feature maps. (d) Our proposed EfficientFCN with codebook generation and codeword assembly for high-resolution feature upsampling in semantic segmentation.}
\label{fig:framework_img0}
\end{figure}

Semantic segmentation or scene parsing is the task of assigning one of the pre-defined class 
labels to each pixel of an input image. It is a fundamental yet challenging task in computer vision.
% and is widely used in many practical applications, \eg, robot sensing, autonomous driving, video surveillance, image editing, etc. 
%In recent years, Deep Convolutional Neural Networks (DCNNs) have dominated many computer vision tasks, including image classification, face recognition, object detection, semantic segmentation, pose estimation, keypoint detection, depth estimation, etc., because of their powerful capability on fitting large scale training data with end-to-end learned feature representations.
The Fully Convolutional Network (FCN) \cite{long2015fully}, as shown in
Fig.~\ref{fig:framework_img0}(a), for the first time demonstrates the success of exploiting a fully convolutional network in semantic segmentation, which adopts a DCNN as the feature encoder (\ie, ResNet\cite{he2016deep}) to extract high-level semantic feature maps and then applies a convolution layer to generate the dense prediction. 
For the semantic segmentation, high-resolution feature maps are critical for achieving accurate
segmentation performance since they contain fine-grained structural information to delineate
detailed boundaries of various foreground regions. In addition, due to the lack of large-scale
training data on semantic segmentation, transferring the weights pre-trained on ImageNet can greatly
improve the segmentation performance. Therefore, most state-of-the-art semantic segmentation methods
adopt classification networks as the backbone to take full advantages of ImageNet pre-training. The resolution of feature maps in the original
classification model is reduced with consecutive pooling and strided
convolution operations to learn high-level
feature representations. The output stride of the final feature map is 32
(OS=32), where the fine-grained structural information is discarded. 
Such low-resolution feature maps cannot fully meet the requirements of semantic segmentation where detailed spatial information is needed. 
To tackle this problem, many works exploit dilated convolution (or atrous convolution) to enlarge the receptive field (RF) while maintaining the resolution of high-level feature maps.
State-of-the-art dilatedFCN based
methods\cite{yu2017dilated,chen2017deeplab,Zhang_2018_CVPR,he2019adaptive,Zhang_2019_CVPR} (shown in
Fig.~\ref{fig:framework_img0}(b)) have demonstrated that removing the downsampling
operation and replacing convolution with the dilated convolution
in the later blocks can achieve superior performance, resulting in 
final feature maps of output stride 8 (OS=8). Despite the superior performance and no extra parameters introduced by dilated convolution,  the high-resolution feature representations require high computational complexity and memory consumption. 
For instance, for an input image with 512$\times$512 and the ResNet101 as the backbone encoder, the computational complexity of the encoder increases from 44.6 GFlops to 223.6 GFlops when adopting the dilated convolution with the strides 2 and 4 into the last two blocks.

Alternatively, as shown in Fig.~\ref{fig:framework_img0}(c), the encoder-decoder based methods (
\eg\ \cite{ronneberger2015u}) exploit using a decoder to gradually upsample and generate the high-resolution feature maps by aggregating multi-level feature representations from the backbone (or the encoder). These encoder-decoder based methods can obtain high-resolution
feature representations efficiently. However, on one hand, the fine-grained structural details are already lost in the topmost high-level feature maps of OS=32. Even with the skip connections, lower-level high-resolution feature maps cannot provide abstractive enough features for achieving high-performance segmentation. 
On the other hand, existing decoders mainly utilize the bilinear upsampling or deconvolution operations to increase the resolution of the high-level feature maps. These operations are conducted in a
local manner. The feature vector at each location of the upsampled
feature maps is recovered from a limited receptive filed. Thus, although the encoder-decoder models are generally faster and more memory friendly than dilatedFCN based methods, their performances generally cannot compete with those of the dilatedFCN models.

To tackle the challenges in both types of models, we propose the EfficienFCN
(as shown in Fig.~\ref{fig:framework_img0}(d)) with the Holistically-guided Decoder (HGD) to bridge the
gap between the dilatedFCN based methods and the encoder-decoder based methods. Our network can adopt any widely
used classification model without dilated convolution as the encoder (such as ResNet models) to
generate low-resolution high-level feature maps (OS=8). Such an encoder is both computationally and
memory efficient than those in DilatedFCN model.
%To verify the effectiveness of our proposed Holistically-guided decoder, we choose the widely used ResNet in semantic segmentation as our encoder. In the proposed Decoder, as shown in Fig.~\ref{fig:framework_our0} we only recover the high-resolution feature representations at the size OS=8 for both the efficiency and giving a fair comparisons with the  dilatedFCN based methods. Further, our proposed HGD can be easily extended to  reconstruct the high resolution map to the larger size.
Given the multi-level feature maps from the last three blocks of the encoder, the proposed holistically-guided decoder takes the advantages of both high-level but low-resolution (OS=32) and also mid-level high-resolution feature maps (OS=8, OS=16) for achieving high-level feature upsampling with semantic-rich features. Intuitively, the higher-resolution feature maps contain more fine-grained structural information, which is beneficial for spatially guiding the feature upsampling process; the lower-resolution feature maps contain more high-level semantic information, which are more suitable to encode the global context effectively. Our HGD therefore generates a series of holistic codewords in a codebook to summarize different global and high-level aspects of the input image from the low-resolution feature maps (OS=32). Those codewords can be properly assembled in a high-resolution grid to form the upsampled feature maps with rich semantic information. Following this principle, the HGD generates assembly coefficients from the mid-level high-resolution feature maps (OS=8, OS=16) to guide the linear assembly of the holistic codewords at each high-resolution spatial location to achieve feature upsampling.
%%Our proposed enjoys superior performance and efficiency at the same time. In our HGD work, we first reorganize the feature maps from the last three maps as two streams of feature maps with size OS=8 and OS=32
%respectively. Intuitively, the higher resolution feature maps contain more fine-grained structural information, which is benefit for the details recovery in the reconstructed high resolution feature maps. And the lower resolution feature maps include more high level semantic information, which fits to encoding the global contextual information efficiently. So we propose to utilize the codebook learning mechanism to learn the semantic codewords from global context information in 
%the low resolution stream.  The we predict the corresponding query coefficient for each location at the high resolution OS=8 from the high resolution stream. Finally, with the semantic codebook and query coefficients, we can efficiently produce the finally high resolution feature maps. Unlike previous decoders that reconstruct the high resolution feature maps in a locally manner via a series of bilinear
%upsampling (or deconvolution) and skip-connection operations, our proposed decoder aims to compute each feature at the high resolution feature maps in a globally manner by a linear combinations of the globally learnt semantic codewords. Moreover, thanks to our well-designed mechanism, our proposed decoder is flexible and can be combined with any classification model (as the encoder) for semantic segmentation. 
Our proposed EfficientFCN with holistically-guided decoder achieves high  segmentation accuracy on three popular public benchmarks, which demonstrate the efficiency and effectiveness of our proposed decoder.

In summary, our contributions are as follows.
\begin{itemize}
    \item We propose a novel holistically-guided decoder, which can efficiently generate the high-resolution feature maps considering holistic contexts of the input image.
   
\item Because of the light weight and high performance of the proposed holistically-guided decoder, our EfficientFCN can adopt the encoder without any dilated convolution but still achieve superior performance.   
        
   \item Our EfficientFCN achieves competitive (or better) results compared
       with the state-of-the-art dilatedFCN based methods on the PASCAL
       Context, PASCAL VOC, ADE20K datasets, with 1/3 fewer FLOPS.
\end{itemize}

\section{Related Work}
In this section, we review recent FCN-based methods for semantic segmentation. Since the successful demonstration of FCN \cite{long2015fully} on semantic segmentation, many methods were proposed to improve the performance the FCN-based methods, which mainly include two categories of methods: the dilatedFCN-based methods and the encoder-decoder architectures.

\noindent \textbf{DilatedFCN.} 
The Deeplab V2 \cite{chen2017deeplab,chen2017rethinking} proposed to exploit dilated convolution in
the backbone to learn a high-resolution feature map, which increases the
output stride from 32 to 8. However, the dilated convolution in the last two layers of the backbone adds huge extra computation and leaves large
memory footprint. Based on the dilated convolution backbone, many works \cite{Zhang_2019_CVPR,fu2019dual,Fu_2019_ICCV,he2019dynamic} continued to apply different strategies as the segmentation heads to acquire the context-enhanced feature maps. PSPNet \cite{zhao2017pyramid} utilized the Spatial Pyramid Pooling (SPP) module to increase the receptive field. EncNet \cite{Zhang_2018_CVPR} proposed an encoding layer to predict a feature re-weighting vector from the global context and selectively high-lights 
class-dependent feature maps. 
CFNet \cite{Zhang_2019_CVPR} exploited an aggregated co-occurrent feature (ACF) module to
aggregate the co-occurrent context by the pair-wise similarities in the feature space.
Gated-SCNN\cite{takikawa2019gated} proposed to use a new gating mechanism to connect the intermediate layers and a new loss function that exploits the duality
between the tasks of semantic segmentation and semantic
boundary prediction.
DANet \cite{fu2019dual} proposed to use two attention modules with the self-attention mechanism to aggregate features from
spatial and channel dimensions respectively.
ACNet \cite{Fu_2019_ICCV} applied a dilated ResNet as the backbone and combined the encoder-decoder strategy for the 
observation that the global context from high-level features
helps the categorization of some large semantic confused
regions, while the local context from lower-level visual features helps to generate sharp boundaries or clear details.
DMNet \cite{he2019dynamic} generated a set of dynamic filters of different
sizes from the multi-scale neighborhoods for handling the scale variations of objects for semantic segmentation.
Although these works further improve the performances on different benchmarks, these proposed heads 
still adds extra computational costs to the already burdensome encoder.

%which leads the dilated based methods lack the efficiency.

\noindent \textbf{Encoder-Decoder.} 
Another type of methods focus on efficiently acquire the high-resolution semantic feature maps via the encoder-decoder architectures. Through
the upsampling operations and the skip connections, the encoder-decoder
architecture \cite{ronneberger2015u} can gradually recover the high-resolution feature maps for 
segmentation. DUsampling \cite{tian2019decoders} designed a data-dependent upsampling module based on fully 
connected layers for constructing the high-resolution feature maps from the low-resolution 
feature maps. FastFCN \cite{wu2019fastfcn} proposed a Joint Pyramid Upsampling (JPU) method via multiple 
dilated convolution to generate the high-resolution feature maps.
One common drawback of these methods is that the feature at each location of the upsampled high-resolution feature maps is 
constructed via only local feature fusion. Such a property limited limits their performance in 
semantic segmentation, where global context is important for the final performance.

\section{Proposed Method}
In this section, We firstly give a thorough analysis of the classical
encoder-decoder based methods. 
Then, to tackle the challenges in the classical encoder-decoder methods, 
we propose the EfficientFCN, which is based on the traditional ResNet as the
encoder backbone 
network for semantic segmentation. In our EfficientFCN, the
Holistically-guided Decoder (HGD)
is designed to recover the high-resolution (OS=8) feature maps from three
feature maps in the last three blocks from the ResNet encoder backbone.
\begin{figure}[tb]
    \centering
    \subfloat{\includegraphics[width=11.0cm]{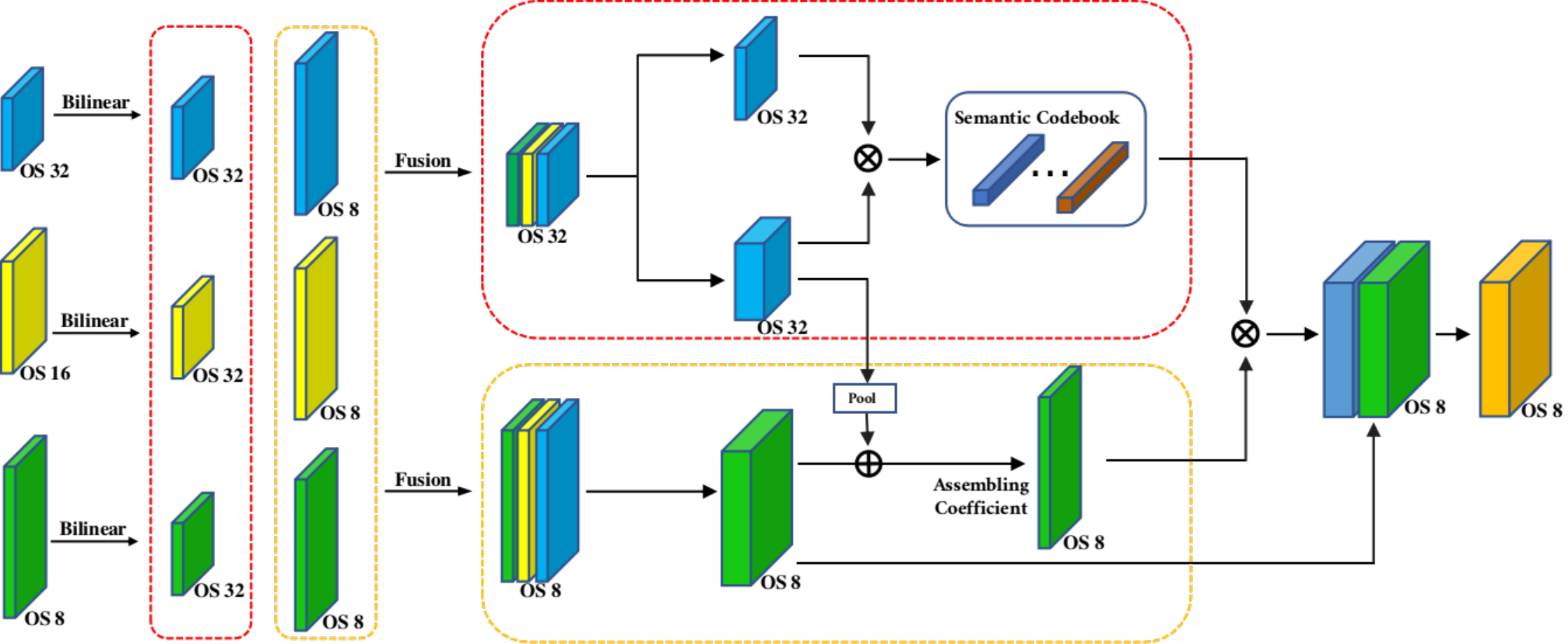}}
\caption{Illustration of the proposed EfficientFCN model. It consists of three main components. Multi-scale feature fusion fuses multi-scale features to obtain OS=8 and OS=32 multi-scale feature maps. Holistic codebook generation results in a series of holistic codewords summarizing different aspects of the global context. High-resolution feature upsampling can be achieved by codeword assembly.}
\label{fig:framework_img1}
\end{figure}
\subsection{Overview}
In state-of-the-art DCNN backbones, the high-resolution mid-level feature maps
at earlier stages can better encode fine-grained structures, while the
low-resolution high-level features at the later stages are generally more
discriminative for category prediction. 
They are essential and complementary information sources for achieving
accurate semantic segmentation.
To combine the strengths of both features, encoder-decoder structures were
developed to reconstruct high-resolution feature maps that have semantic-rich
information from the multi-scale feature maps. 
To generate the final feature map $\tilde{f}_8$ of OS=8 for mask prediction, a
conventional three-stage encoder-decoder first upsamples the deepest encoder
feature map $f_{32}$ of OS=32 to generate OS=16 feature maps $f_{16}$. The
OS=16 feature maps $e_{16}$ of the same size from the encoder is either
directly concatenated as $[f_{16}; e_{16}]$ (U-Net \cite{ronneberger2015u}) or summed as
$f_{16}+e_{16}$ followed by some $1\times 1$ convolution to
generate the upsampled OS=16 feature maps, $\tilde{f}_{16}$. The same
upsampling + skip connection procedure repeats again for $\tilde{f}_{16}$ to
generate $\tilde{f}_{8}$. The upsampled features $\tilde{f}_8$ contain both
mid-level and high-level information to some extent and can be used to
generate the segmentation masks.
%can be described as  
%\begin{align}
%    z &= F_2(x_1, G_2(F_1(x_2, G_1(x_3)))) 
%\end{align}
%where $z$ is the reconstructed feature map with OS = 8, $x_1, x_2, x_3$ are
%the feature maps with OS=8, OS=16, OS=32 from the last three stages of the
%encoder backbone (\eg, a ResNet-50), respectively. $G_1$ and $G_2$
%denote the upsampling operations, which can be modeled as bilinear upsampling
%or deconvolution operations. $F_1$ and $F_2$ can be a series of $3\times 3$
%convolution layers. NO NEED TO HAVE EQUATION 1 AT ALL.
However, since bilinear upsampling and deconvolution layer in the classical decoder are local operations
with limited receptive field. We argue that they are incapable of exploring
important global context of the input image, which are crucial for achieving
accurate segmentation. 
Although there were some existing attempts \cite{Zhang_2018_CVPR,hu2018squeeze} on using
global context to re-weight the contributions of different channels of the
feature maps either in the backbone \cite{Zhang_2018_CVPR} or in the upsampled feature
maps \cite{hu2018squeeze}. This strategy only scales each feature channel but
maintains the original spatial size and structures. Therefore, it is incapable
of generating high-resolution semantic-rich feature maps to improve the
recovery of fine-grained structures.
%, each feature $z_i$ is computed at three small patches from the feature maps
%$x_1$, $x_2$ and $x_3$.  This property of the classical encoder-decoder
%methods limits their performance for the semantic segmentation as this
%locally-limited feature fusion method cannot promise to recovery the high
%quality semantic information.
To solve this drawback, we propose a novel Holistically-guided Decoder (HGD),
which decomposes the feature upsampling task into the generation of a series
of holistic codewords from high-level feature maps to capture global contexts,
and linearly assembling codewords at each spatial location for semantic-rich
feature upsampling. Such a decoder can exploit the global contextual information
to effectively guide the feature upsampling process and is able to recover
fine-grained details. 
%has the ability to generate a high resolution feature maps with the holistic
%context information from the feature maps $x_1, x_2, x_3$. 
Based on the proposed HGD, we present the EfficientFCN (as shown in
Fig.~\ref{fig:framework_img0}(d)) with an efficient encoder free of dilated
convolution for efficient semantic segmentation. 
\subsection{Holistically-guided Decoder for Semantic-rich Feature Upsampling}
To take advantages of both low-resolution high-level feature maps of size
OS=32 and the high-resolution mid-level feature maps of sizes OS=8 and OS=16,
since the high-level feature maps have already lost most structural details
but are semantic-rich to encode categorical information, we argue that
recovering detailed structures from them is quite challenging and also
non-necessary. Instead, we propose to generate a series of holistic codewords
without any spatial order from the high-level feature maps to capture
different aspects of the global context. On the other hand, the mid-level
high-resolution feature maps have maintained abundant image structural
information. But they are from relatively shallower layers and cannot encode
accurate enough categorical features  for final mask prediction. However, they
would still be representative enough for guiding the linear assembly of the
semantic-rich codewords for high-resolution feature upsampling. Our proposed
Holistically-guided Decoder therefore contains three main components:
multi-scale feature fusion, holistic codebook generation, and codeword
assembly for high-resolution feature upsampling.
%For the feature maps from last three stages of the ResNet encoder backbone,
%we firstly use a
%$1\times1$ convolution layer for each feature maps to compress its channel to
%512 for the lower computational complexity, then acquire the feature maps
%$x_1$, $x_2$ and $x_3$.
%$x_1$ is at the higher resolution with OS = 8 and contains more fine-grained
%structural 
%information. $x_2$ (OS=16) and $x_3$ (OS=32) have the lower resolution but
%encoded much richer semantic information. 
%So our proposed HED utilizes the holistic context information in the
%multi-scale feature maps with the aim of learning the powerful semantic
%information from them and applying the guidance from $x_1$ to generate
%a high resolution feature map with the powerful semantic context information.
%To construct the high resolution feature maps from the holistic context
%information with a small computational consumption, our proposed HED mainly
%consists
%of three parts: multi-scale features shuffle, semantic codebook learning,
%query feature learning. Among them, the semantic codebook learning step
%predicts the semantic codewords from the holistic context at the feature maps
%at the resolution OS=32 with the aim of reducing the computational
%complexity. 
%The query feature learning step learns the query coefficient for each
%location at the 
%output feature maps with the resolution OS=8.  

%\vspace{4pt}
\noindent \textbf{Multi-scale features fusion.}
Given the multi-scale feature maps from the encoder, although we can directly
encode the holistic codewords from the high-level OS=32 feature maps and also
directly generate the codeword combination coefficients from the mid-level
OS=16 and OS=8 feature maps, we observe the fusion of multi-scale feature maps
generally result in better performance. 
For the OS=8, OS=16, OS=32 feature maps from the encoder, we first adopt
separate $1\times 1$ convolutions to compress each of their channels to 512
for reducing the follow-up computational complexity, obtaining $e_{8}, e_{16},
e_{32}$, respectively. The multi-scale fused OS=32 feature maps $m_{32}$ are
then obtained by downsampling $e_{8}, e_{16}$ to the size of OS=32 and
concatenating them along the channel dimension with $e_{32}$ as $m_{32} = [
    e_{8}^\downarrow;  e_{16}^\downarrow; e_{32}] \in \mathbb{R}^{1536\times
(H/32) \times (W/32)}$, where $^\downarrow$ represents bilinear downsampling,
$[\cdot;\cdot]$ denotes concatenation along the channel dimension, and $H$ and
$W$ are the input image's height and width, respectively. We can also obtain
the multi-scale fused OS=8 feature maps, $m_8 = [ e_{8}; e_{16}^\uparrow;
e_{32}^\uparrow] \in \mathbb{R}^{1536\times (H/8) \times (W/8)}$, in a similar
manner.
%Although we can directly conduct the semantic codebook learning on the
%original
%$x_3$ and predict the query feature from the original $x_1$, we find that the
%combination of these different scales feature maps are both beneficial for
%the
%effectiveness of the semantic codebook learning and query feature learning. 
%In our proposed HED, we use this multi-scale features shuffle strategy to
%fusion these feature maps. As shown in Fig.~\ref{fig:framework_img1_b},
%the multi-scale features shuffle step can be described as:
%\begin{align}
%    s_1 = \text{Concat}(x_1, S_1(x_2), S_1(x_3)),\
%    s_2=\text{Concat}(S_2(x_1),S_2(x_2), x_3)
%\end{align}
%where $S_1$ is the bilinear downsampling operation, $S_2$ is the bilinear
%upsampling operation. $s_1$ and $s_2$ are the feature maps with resolution
%OS=32 and OS=8, respectively.

%\vspace{4pt}
\noindent \textbf{Holistic codebook generation.}
Although the mutli-scale fused feature maps $m_{32}$ are created to integrate
both high-level and mid-level features, their small resolutions make them lose
many structural details of the scene. On the other hand, because $e_{32}$ is
encoded from the deepest layer, $m_{32}$ is able to encode rich categorical
representations of the image. We therefore propose to generate a series of
unordered holistic codewords from $m_{32}$ to implicitly model different
aspects of the global context.
To generate $n$ holistic codewords, a codeword base map $B \in
\mathbb{R}^{1024 \times (H/32)\times}$ $^{(W/32)}$ and $n$ spatial weighting
maps $A \in \mathbb{R}^{n \times (H/32)\times (W/32)}$ are first computed from
the fused multi-scale feature maps $m_{32}$ by two separate $1\times 1$
convolutions. 
For the bases map $B$, we denote $B(x,y) \in \mathbb{R}^{1024}$ as the 1024-d
feature vector at location $(x,y)$; for the spatial weighting maps $A$, we use
$A_i \in \mathbb{R}^{(H/32) \times (W/32)}$ to denote the $i$th weighting map.
To ensure the weighting maps $A$ are properly normalized, the softmax function is adopted to 
normalize all spatial locations of each channel $i$ (the $i$-th spatial feature map) as
\begin{align}
            \tilde{A}_i(x,y) = \frac{\exp(A_i(x,y))}{\sum_{p,q}
            \exp(A_i(p,q))}.
\end{align}
The $i$-th codeword $c_i\in \mathbb{R}^{1024}$ can be obtained as the weighted
average of all codeword bases $B(x,y)$, \ie,
\begin{align}
            c_i = \sum_{p,q} \tilde{A}_i(p,q) B(p,q).
\end{align}
In other words, each spatial weighting map $\tilde{A}_i$ learns to linearly
combine all codeword bases $B(x,y)$ from all spatial locations to form a
single codeword, which captures certain aspect of the global context. The $n$
weighting maps eventually result in $n$ holistic codewords $C = [c_1, \cdots,
c_n] \in \mathbb{R}^{1024 \times n}$ to encode high-level global features.
%To generate a holistically-guided high resolution context, one efficient
%method is to predict some class-agnostic semantic codewords from the holistic
%context and use these predicted semantic codewords to construct the output
%high resolution feature map. In our HED, we apply an attention mechanism to
%learn the semantic codebook from the feature map $s_1$. To generate a
%semantic
%codebook with $n$ semantic codewords, we use one
%$1\times 1$ convolution block to generate the feature map $a$ with channel
%$n$, and one $1\times 1$ convolution block to generate feature map $v$ with
%channel 1024. Then, with the feature maps $a$ and $v$, each semantic word
%$c_i$
%in the semantic $C=\{c_1,\dots,c_n\}$ can be computed in a weighted average
%manner:
%\begin{align}
%    c_i = \frac{\sum_{j=1}^{h_1*w_1} a_ij * v_j}{\sum_{j}^{h_1*w_1} a_ij}
%\end{align}
%where $a_{ij}$ is the value at the location $j$ from the channel $i$ in the
%feature map $a$, $v_j\in\mathbb{R}^{1024}$ is a feature vector at the
%location $j$ at the feature map $v$, $h_1$ and $w_1$ are the spatial size of
%the feature maps $a$ and $v$.
%
%In this manner, each learnt semantic codeword $c_i$ is computed from the
%holistic
%context and can be used for the holistically-guided decoding operation.
%Moreover, to make sure the semantic codewords be more effective to represent
%the
%semantic object parts or attributes, we add a semantic loss on feature map
%$a$. For feature map $a$, we use a $1\times 1$ convolution layer to get a
%prediction output with the size OS=32, and compute a semantic segmentation
%loss as one auxiliary loss.

%\vspace{4pt}
\noindent \textbf{Codeword assembly for high-resolution feature upsampling.} 
The holistic codewords can capture various global contexts of the input image.
They are perfect ingredients for reconstructing the high-resolution
semantic-rich feature maps as they are encoded from the high-level features
$m_{32}$. However, since their structural information have  been mostly
removed during codeword encoding, we turn to use the OS=8 multi-scale fused
features $m_8$ to predict the linear assembly coefficients of the $n$
codewords at each spatial location for creating a high-resolution feature map.
More specifically, we first create a raw codeword assembly guidance feature
map $G \in \mathbb{R}^{1024\times (H/8)\times (W/8)}$ to predict the assembly
coefficients at each spatial location, which are obtained by applying a
$1\times 1$ convolution on the multi-scale fused features $m_8$. However, the
OS=8 fused features $m_8$ have no information on the holistic codewords as they
are all generated from $m_{32}$. We therefore consider the general codeword
information as the global average vector of the codeword based map $\bar{B} \in
\mathbb{R}^{1024}$ and location-wisely add it to the raw assembly guidance
feature map to obtain the novel guidance feature map $\bar{G} = G \oplus \bar{B}$, where $\oplus$ represents
location-wise addition. Another $1\times 1$ convolution applied on the
guidance feature map $\bar{G}$ generates the linear assembly weights of the $n$
codewords $W \in \mathbb{R}^{n \times (H/8) \times (W/8)}$ for all $(H/8)
\times (W/8)$ spatial locations. By reshaping the weighting map $W$ as an $n
\times (HW/8^2)$ matrix, the holistically-guided upsampled feature
$\tilde{f}_8$ can be easily obtained as
\begin{align}
            \tilde{f}_8 = W^\top C.
\end{align}
Given the holistically-guided upsampled feature map $\tilde{f}_8$, we reconstruct the final 
upsampled feature map  $\hat{f}_8$ by concatenating the feature map $\tilde{f}_8$ with the
guidance feature map $G$. 
Such an upsampled feature map $\hat{f}_8$ takes advantages of both $m_8$
and $m_{32}$, and contains semantic-rich and also structure-preserved features
for achieving accurate segmentation. 

%\vspace{4pt}
\noindent {\bf Final segmentation mask.} Given the upsampled feature map
$\hat{f}_8$, a $1\times 1$ convolution can output a segmentation map of
OS=8, which is further upsampled back to the original resolution $H
\times W$ as the final segmentation mask.

%%%%%%%%%%%%%%%%%%%%%%%%%%%%%%%%%%%%%%%%%%%%%%

%%%%%%%%%%%%%%%%%%%%%%%%%%%%%%%%%%%%%%%%%%%%%%

\section{Experiments}
In this section, we introduce the implementation details, training strategies and evaluation metrics of our experiments.
To evaluate our proposed EfficientFCN model, we conduct comprehensive experiments
on three public datasets PASCAL Context \cite{mottaghi2014role}, PASCAL VOC 2012 \cite{everingham2010pascal}
and ADE20K \cite{zhou2017scene}. To further evaluate the contributions of individual components in our model, we conduct detailed ablation studies on the PASCAL Context dataset. 
%
%Finally, we report our results and provide comparisons with state-of-the-art methods on PASCAL Context, PASCAL VOC 2012 and ADE20K.
\subsection{Implementation Details}
\noindent
\textbf{Network Structure.}\
Different with the dilatedFCN based methods, which remove the stride of the last two blocks of
the backbone networks and adopt the dilated convolution with the dilation rates $2$ and $4$,
we use the original ResNet \cite{he2016deep} as our encoder backbone network. 
Thus the size of the output feature maps from the last ResBlock is $32\times$ smaller than that of
the input image. After feeding the encoder feature maps into our proposed holistic-guided decoder,
the classification is performed on the output upsampled feature map $\hat{f}_8$. 
%The final segmentation mask obtained by upscaling the results to the size of input image via bilinear interpolation.
The ImageNet \cite{russakovsky2015imagenet} pre-trained weights are utilized to initialize the encoder network.
%
%All experiments are implemented with PyTorch \cite{paszke2017automatic}.
%
%

%\vspace{4pt}
\noindent
\textbf{Training Setting.}\
A poly learning rate policy \cite{chen2017deeplab} is used in our experiments. We set the initial learning rates as $0.001$
for PASCAL Context \cite{mottaghi2014role}, $0.002$ for PASCAL VOC 2012
\cite{everingham2010pascal} and 
ADE20K \cite{zhou2017scene}. The power of poly learning rate policy is set as $0.9$. The optimizer
is stochastic gradient descent (SGD) \cite{bottou2010large} with momentum $0.9$ and weight
decay $0.0001$.
We train our EfficientFCN for $120$ epochs on PASCAL Context, $80$ epochs on PASCAL 2012 and $120$ epochs on ADE20K, respectively. We set the crop size to $512\times 512$ on PASCAL Context and PASCAL 2012. Since the average image size is larger than other two datasets, we use $576 \times 576$ as the crop size on ADE20K. For data augmentation, we only randomly flip the input image and scale it randomly in the range $[0.5, 2.0]$.
%
%

%\vspace{4pt}
\noindent
\textbf{Evaluation Metrics.}\
We choose the standard evaluation metrics of pixel accuracy (pixAcc) and mean Intersection of Union (mIoU) as the evaluation metrics in our experiments. Following the best practice \cite{Zhang_2018_CVPR,he2019adaptive,fu2019dual}, 
we apply the strategy of averaging the network predictions in multiple scales for evaluation. For
each input image, we first randomly resize the input image with a scaling factor sampled uniformly
from [0.5, 2.0] and also randomly horizontally flip the image. These predictions are then averaged to generate the final prediction.
\begin{table}[tb]
\centering
%\hspace{10pt}
%\begin{minipage}[t]{0.43\textwidth}
%
%\begin{table}[tb]
%\small
\centering
\caption{Comparisons with classical encoder-decoder methods.}
\label{table:ablation_encoder_decoder}
%\resizebox{0.5\columnwidth}{!}{
\setlength{\tabcolsep}{0.5mm}{
\begin{tabular}{llllll}
\hline
\textbf{Method} & \textbf{Backbone} & \textbf{OS} &\textbf{mIoU\%} &
\textbf{Parameters (MB)} &\textbf{GFlops (G)}\\\hline\hline
FCN-32s  &  ResNet101 & 32 & 43.3 & 54.0 & 44.6 \\ %\cline{1-1} \cline{3-3}
dilatedFCN-8s  & dilated-ResNet101 & 8 & 47.2 &54.0 & 223.6 \\ %\cline{1-1} \cline{3-3}
UNet-Bilinear  & ResNet101 & 8 & 49.3 & 60.7 & 87.9 \\ %\cline{1-1} \cline{3-3}
UNet-Deconv  & ResNet101 & 8  & 49.1 & 62.8 & 93.2\\ %\cline{1-1} \cline{3-3}
\hline
EfficientFCN & ResNet101 & 8  & 55.3 & 55.8 & 69.6\\ %\cline{1-1} \cline{3-3}
\hline
\end{tabular}}
%\end{minipage}
\end{table}
\subsection{Results on PASCAL Context}
The PASCAL Context dataset consists of 4,998 training images and 5,105 testing images for scene parsing. It is a complex and challenging dataset based on PASCAL VOC 2010 with more annotations and fine-grained scene classes, which includes 59 foreground classes and one background class. 
We take the same experimental settings and evaluation strategies following previous works \cite{Zhang_2018_CVPR,Zhang_2019_CVPR,fu2019dual,Fu_2019_ICCV,he2019dynamic}.
We first conduct ablation studies on this dataset to demonstrate the
effectiveness of each individual module design of our proposed EfficientFCN
and then compare our model with state-of-the-art methods. The ablation studies are conducted with a ResNet101 encoder backbone.
\begin{table}[tb]
\centering
\begin{minipage}[t]{0.44\textwidth}
%
%\begin{table}[tb]
%\small
%\centering
\caption{Results of using different numbers of scales for multi-scale fused feature $m_{32}$ to generate the holistic codewords.}
\label{table:ablation_ms_codebook}
%\resizebox{\columnwidth}{!}{
\setlength{\tabcolsep}{1mm}{
\begin{tabular}{c|cccc}
\hline
        {} & {$\{32\}$} & {$\{16, 32\}$} & {$\{8, 16, 32\}$} \\ \hline\hline
        pixAcc &80.0 & 80.1 & 80.3  \\
        mIoU & 54.8 & 55.1 & 55.3   \\
%ResNet101 & \textbf{\checkmark} & \textbf{\checkmark} & \textbf{\checkmark} &  \textbf{\checkmark} &\textbf{82.82} \\
\hline
\end{tabular}}
\end{minipage}
%\hfill
\begin{minipage}[t]{0.11\textwidth}
\end{minipage}
\begin{minipage}[t]{0.44\textwidth}
%
%\begin{table}[tb]
%\small
\centering
\caption{Results of using different numbers of scales for multi-scale fused feature $m_8$ to estimate codeword assembly coefficients.}
\label{table:ablation_ms_assembly}
%\resizebox{\columnwidth}{!}{
\setlength{\tabcolsep}{1mm}{
\begin{tabular}{c|cccc}
\hline
        {} & {$\{8\}$} & {$\{8, 16\}$} & {$\{8,16,32\}$} \\ \hline\hline
        pixAcc &78.9 & 80.0 & 80.3  \\
        mIoU & 47.9 & 52.1 & 55.3   \\
%ResNet101 & \textbf{\checkmark} & \textbf{\checkmark} & \textbf{\checkmark} &  \textbf{\checkmark} &\textbf{82.82} \\
\hline
\end{tabular}}
\end{minipage}
\end{table}
%\textbf{FLOPS and memory consumption of different methods.} ...

%\vspace{3pt}
\noindent \textbf{Comparison with the classical encoder-decoders.} For the classical encoder-decoder based methods, the feature upsampling is achieved via either bilinear interpolation or
deconvolution. We implement two classical encoder-decoder based methods, which include the
feature upsampling operation (bilinear upsampling or deconvolution) and the skip-connections. 
To verify the effectiveness of our proposed HGD,  these two methods are trained and tested 
on the PASCAL Context dataset with the same training setting as our model.
The results are shown in Table \ref{table:ablation_encoder_decoder}. Although the classical 
encoder-decoder methods have similar computational complexities, their performances 
are generally far inferior than our EfficientFCN. The key reason is that their upsampled 
feature maps are recovered in a local manner. The simple bilinear interpolation or 
deconvolution cannot effectively upsample the OS=32 feature maps even with the skip-connected 
OS=8 and OS=16 feature maps. In contrast, our proposed HGD can effectively upsample the 
high-resolution semantic-rich feature maps not only based on the fine-grained structural 
information in the OS=8 and OS=16 feature maps but also from the holistic semantic information from the OS=32 feature maps.  

%\vspace{3pt}
\noindent \textbf{Multi-scale features fusion.} We conduct two experiments on multi-scale features fusion to verify their effects on semantic codebook generation and codeword assembly for feature upsampling. In our holistically-guided decoder, the semantic codewords are generated based on the OS=32 multi-scale fused feature maps $m_{32}$ and the codewords assembly coefficients are predicted from the OS=8 multi-scale fused feature maps $m_8$.
For the codeword generation, we conduct three experiments to generate the semantic codebook from
multi-scale fused features with different numbers of scales. As shown in Table
\ref{table:ablation_ms_codebook}, when reducing the number of fusion scales from 3 to 2 and 
from 2 to 1, the performances of our EfficientFCN slightly decrease. The phenomenon is reasonable as the deepest feature maps contain more categorical information than the OS=8 and OS=16 feature maps. 
For the codeword assembly coefficient estimation, the similar experiments are conducted, where results are shown in Table \ref{table:ablation_ms_assembly}. However, different from the above results, when fewer scales of feature maps are used to form the multi-scale fused OS=8 feature map $m_8$, the performances of our EfficientFCN show significant drops.
These results demonstrate that although the OS=8 feature maps contain more fine-grained structural information, the semantic features from higher-level feature maps are essential for guiding the recovery of the semantic-rich high-resolution feature maps.
% deeper stages is essential for a semantic-rich  high-resoltuion feature map recovery. 
%\textbf{Multi-scale feature fusion for generating the coefficients of codewords assembly.}.
\begin{table}[!t]
%\begin{table}[tb]
\centering
%\begin{minipage}[t]{0.43\textwidth}
%
%\begin{table}[tb]
%\small
%\centering
\caption{Ablation study of the number of the semantic codewords.}
\label{table:ablation_n_codewords}
%\resizebox{\columnwidth}{!}{
%\resizebox{0.6\columnwidth}{!}{
%\setlength{\tabcolsep}{7nm}
%\begin{tabular}{c|cccccc}
\setlength{\tabcolsep}{2mm}{
\begin{tabular}{c|cccccc}
\hline
        %{} & \hspace{3pt}{32}\hspace{3pt} & {64} & {128}  & {256} &{512} &{1024}\\ \hline\hline
        {} & {32} & {64} & {128}  & {256} &{512} &{1024}\\ \hline\hline
        pixAcc &79.9 & 80.1 & 80.1 & 80.3 & 80.3 &80.1 \\
        mIoU & 54.5 & 54.9 & 55.0 &55.3 & 55.5 & 55.1  \\
        GFLOPS & 67.9 & 68.1 & 68.6 & 69.6 & 72.1 & 78.9 \\
%ResNet101 & \textbf{\checkmark} & \textbf{\checkmark} & \textbf{\checkmark} &  \textbf{\checkmark} &\textbf{82.82} \\
\hline
\end{tabular}}
%\end{minipage}
\end{table}
%\hfill
\begin{table}[!t]
%\begin{table}[tb]
%\small
\centering
\caption{Segmentation results of state-of-the-art methods on PASCAL Context and ADE20K validation dataset.}
\label{table:pascal-context}
\centering
%\resizebox{\columnwidth}{!}{
\begin{tabular}{lllll}
\hline
    \textbf{Method} & \textbf{Backbone} &\begin{tabular}{c} 
        \textbf{mIoU\%}\\ (PASCAL Context) \end{tabular} & \begin{tabular}{c} 
        \textbf{mIoU\%} \\ (ADE20K) \end{tabular} & \textbf{GFlops} \\\hline\hline
%FCN-8S \cite{long2015fully} &  & 37.8 & 1666 \\ %\cline{1-1} \cline{3-3}
%CRF-RNN \cite{CRF-RNN} &  & 39.3 & \\ %\cline{1-1} \cline{3-3}
%ParseNet \cite{ParseNet} &  & 40.4 &  \\ %\cline{1-1} \cline{3-3}
%BoxSup \cite{BoxSup} &  & 40.5 & \\ %\cline{1-1} \cline{3-3}
%HO\_CRF \cite{HO_CRF} &  & 41.3 & \\ %\cline{1-1} \cline{3-3}
%Piecewise \cite{Piecewise} &  & 43.3 & \\ %\cline{1-1} \cline{3-3}
%VeryDeep \cite{VeryDeep} &  & 44.5&  \\ %\hline
    DeepLab-v2 \cite{chen2017deeplab} &  Dilated-ResNet101-COCO & 45.7 & - & $>$223\\
    RefineNet \cite{RefineNet} &  Dilated-ResNet152 & 47.3 & - & $>$223\\
    MSCI \cite{MSCI} &  Dilated-ResNet152 & 50.3 & - & $>$223\\
%CCL \cite{ccl} & ResNet101 & 51.6 \\
    PSPNet \cite{zhao2017pyramid} & Dilated-ResNet101 & - & 43.29 & $>$223 \\
    SAC \cite{zhang2017scale} & Dilated-ResNet101 & - & 44.30 & $>$223  \\
    EncNet \cite{Zhang_2018_CVPR} &  Dilated-ResNet101 & 51.7 & 44.65 & 234\\
    DANet \cite{fu2019dual} &  Dilated-ResNet101 & 52.6 & - & $>$223 \\ 
    APCNet \cite{he2019adaptive} &  Dilated-ResNet101 & 54.7 & 45.38 & 245\\ 
    CFNet \cite{Zhang_2019_CVPR} &  Dilated-ResNet101 & 54.0 & 44.89 & $>$223 \\ 
    ACNet \cite{Fu_2019_ICCV} & Dilated-ResNet101 & 54.1 & \textbf{45.90} & $>$223\\ 
    APNB \cite{zhu2019asymmetric} & Dilated-ResNet101 & 52.8 & 45.24 & $>$223  \\ 
    DMNet \cite{he2019dynamic} & Dilated-ResNet101 & 54.4 & 45.50 & 242\\
\hline
%Ours & ResNet50 & 52.5 & 1666 \\
    Ours & ResNet101 & \textbf{55.3} & {45.28} & \textbf{70} \\ \hline
\end{tabular}
\end{table}

\begin{figure}[!t]
\centering
%\vspace{-0.1in}
%\begin{comment}
\begin{center}
\begin{tabular}{C{2.2cm}C{2.2cm}C{2.2cm}C{2.2cm}C{2.2cm}}
    \includegraphics[width=2.20cm]{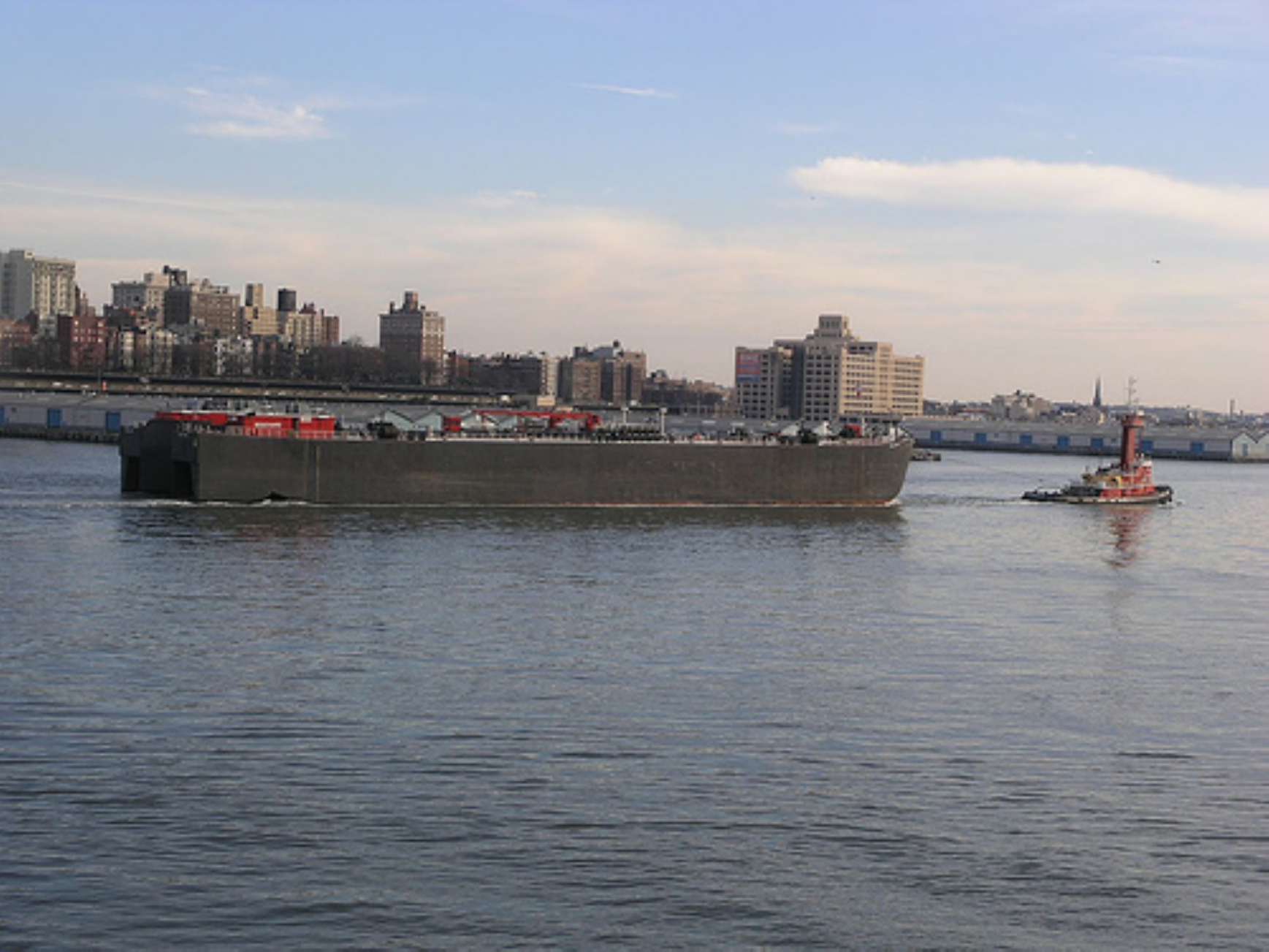} &
    \includegraphics[width=2.20cm]{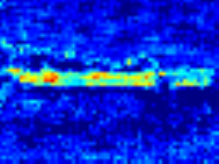} &
    \includegraphics[width=2.20cm]{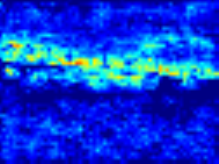} &
    \includegraphics[width=2.20cm]{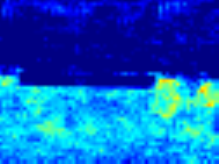} &
    \includegraphics[width=2.20cm]{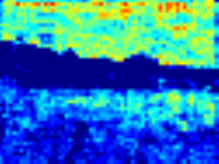} \\
    \includegraphics[width=2.20cm]{figs/exp1_fig1_a} &
    \includegraphics[width=2.20cm]{figs/exp1_fig1_b} &
    \includegraphics[width=2.20cm]{figs/exp1_fig1_c} &
    \includegraphics[width=2.20cm]{figs/exp1_fig1_d} &
    \includegraphics[width=2.20cm]{figs/exp1_fig1_e} \\
    \includegraphics[width=2.20cm]{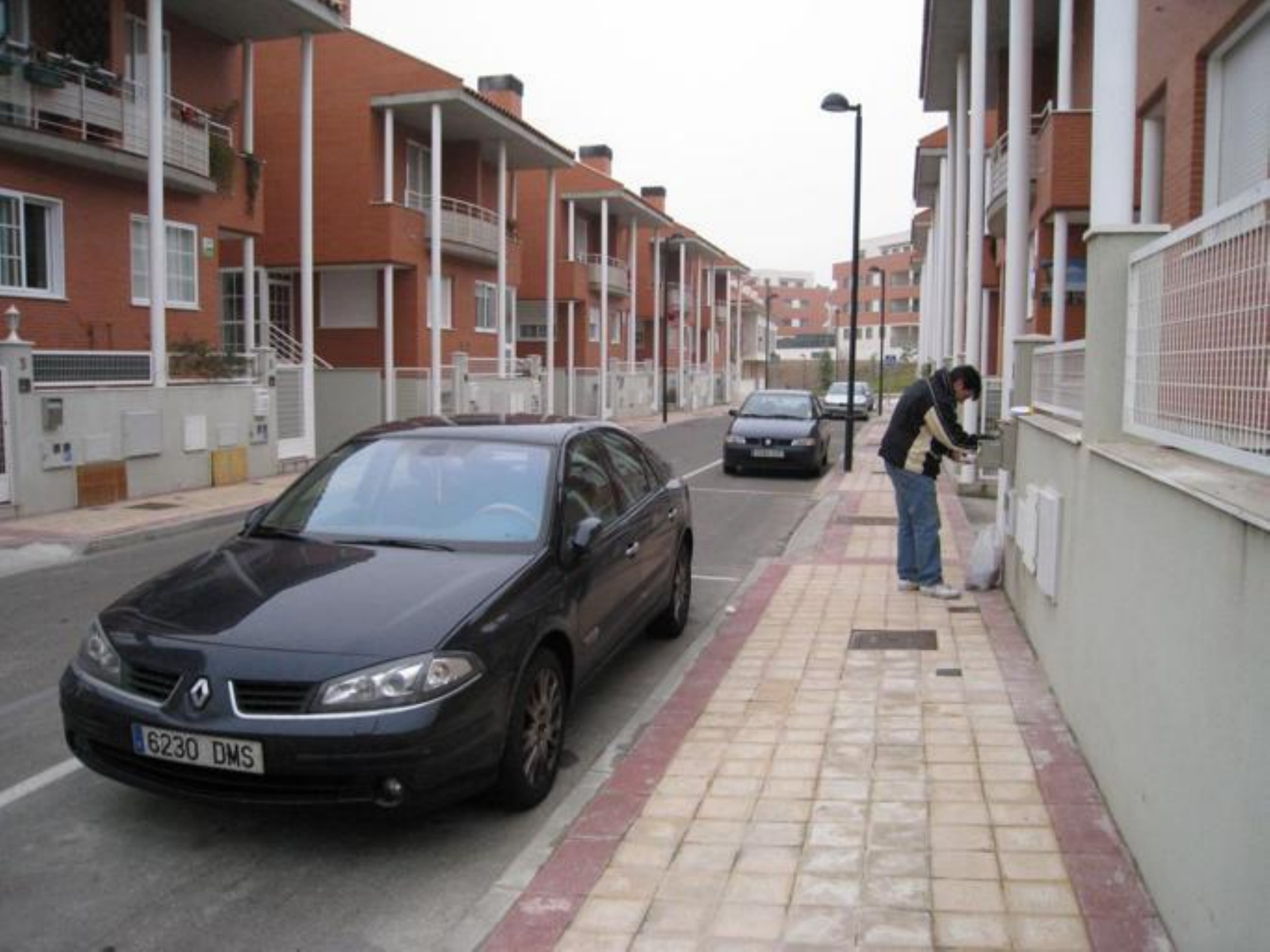} &
    \includegraphics[width=2.20cm]{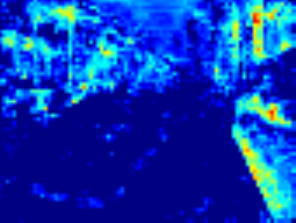} &
    \includegraphics[width=2.20cm]{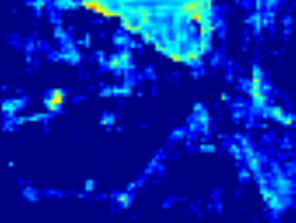} &
    \includegraphics[width=2.20cm]{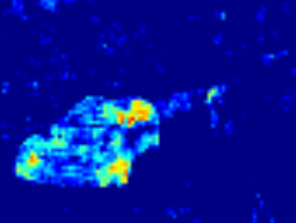} &
    \includegraphics[width=2.20cm]{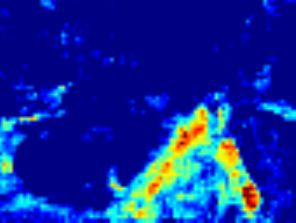} \\
    \centering (a) & (b) & (c) & (d) & (e) \\
\end{tabular}
\end{center}
\caption{(a) Input images from the PASCAL Context and ADE20K dataset. (b-e) Different weighting maps $\tilde{A}_i$ for creating the holistic codewords.}
\label{fig:weighting_maps}
%\end{figure}

%\begin{comment}
%\begin{figure}[tb]
%\centering
\begin{center}
\begin{tabular}{C{2.50cm}C{2.50cm}C{2.50cm}C{2.50cm}}
    \includegraphics[width=2.50cm]{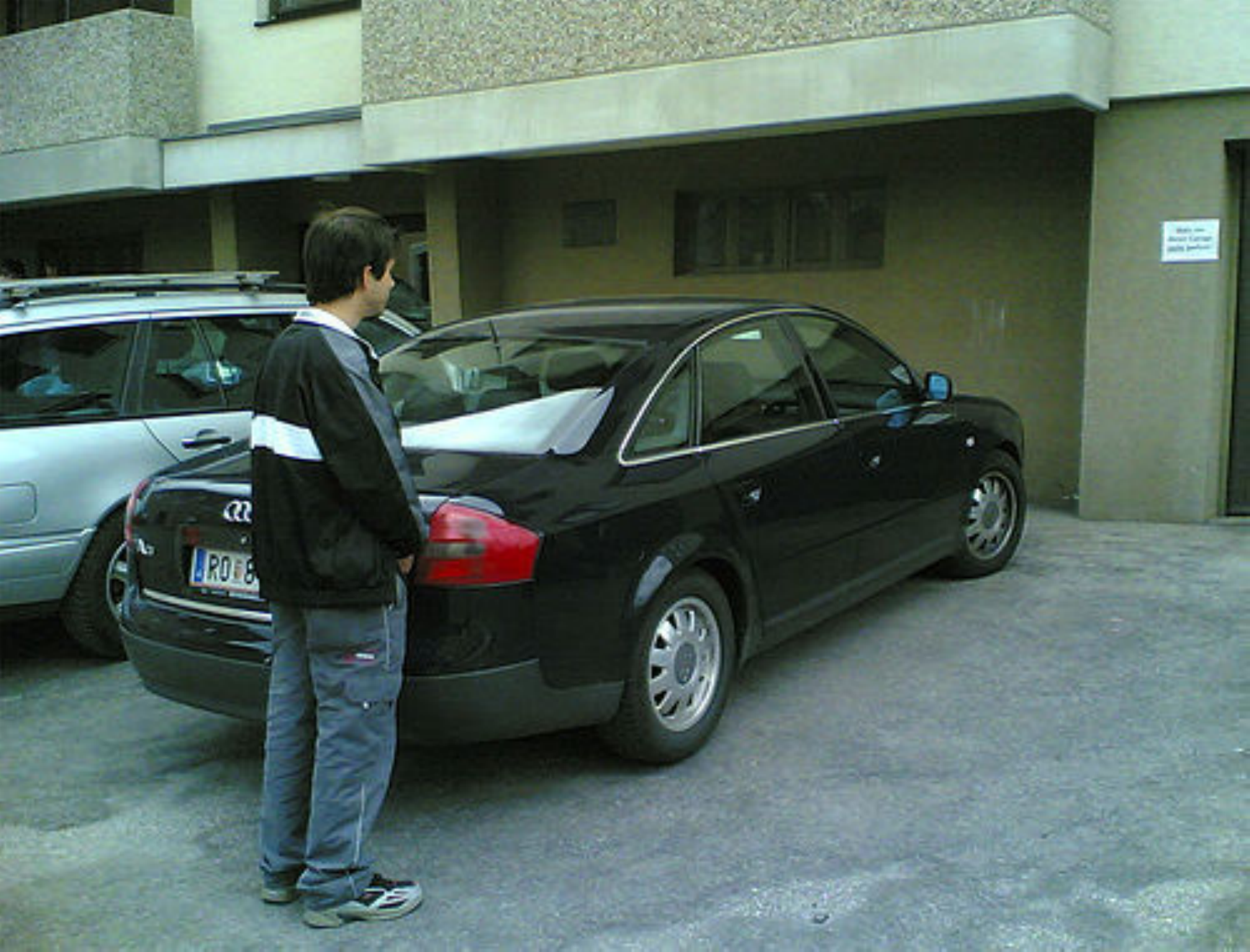} &
    \includegraphics[width=2.50cm]{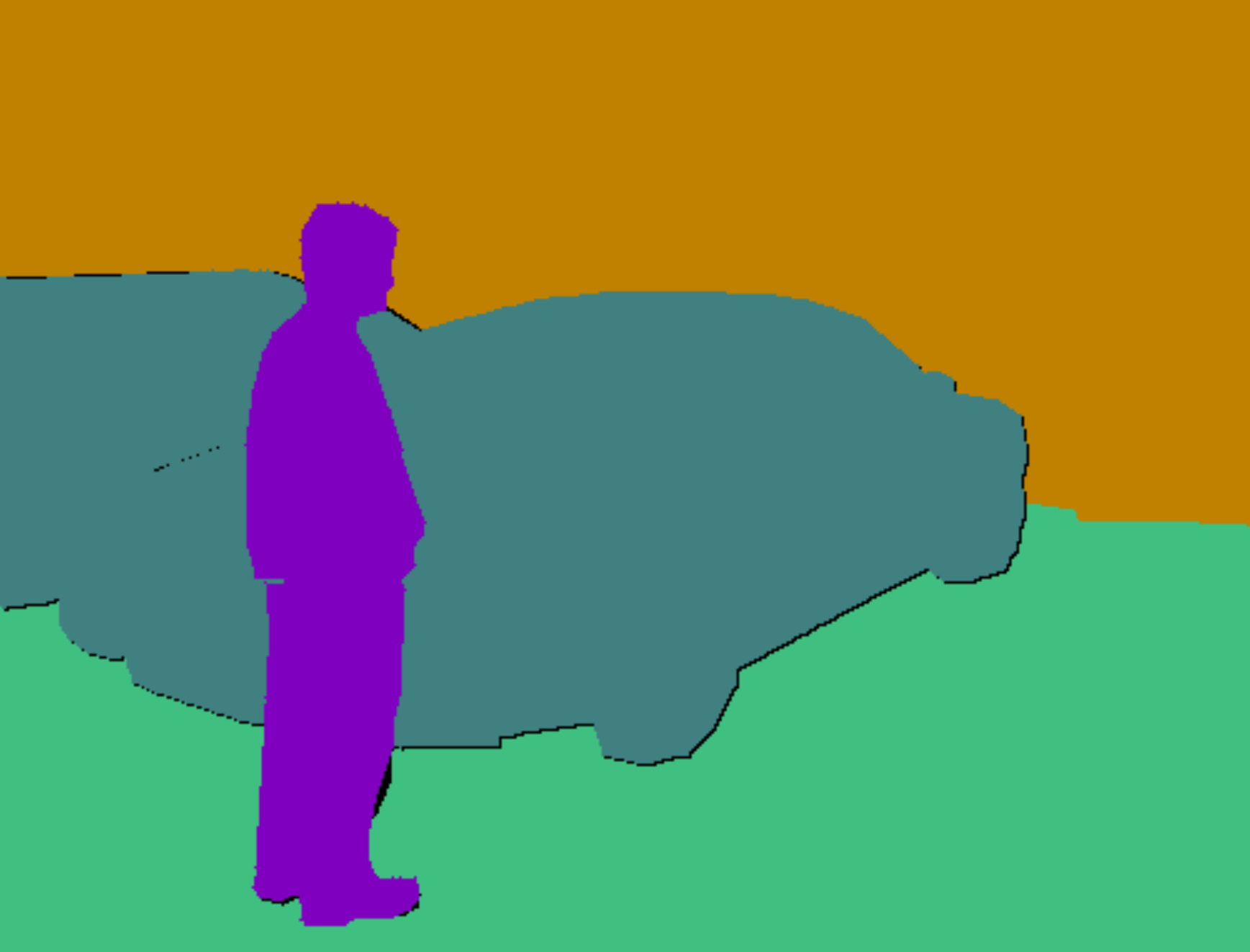} &
    \includegraphics[width=2.50cm]{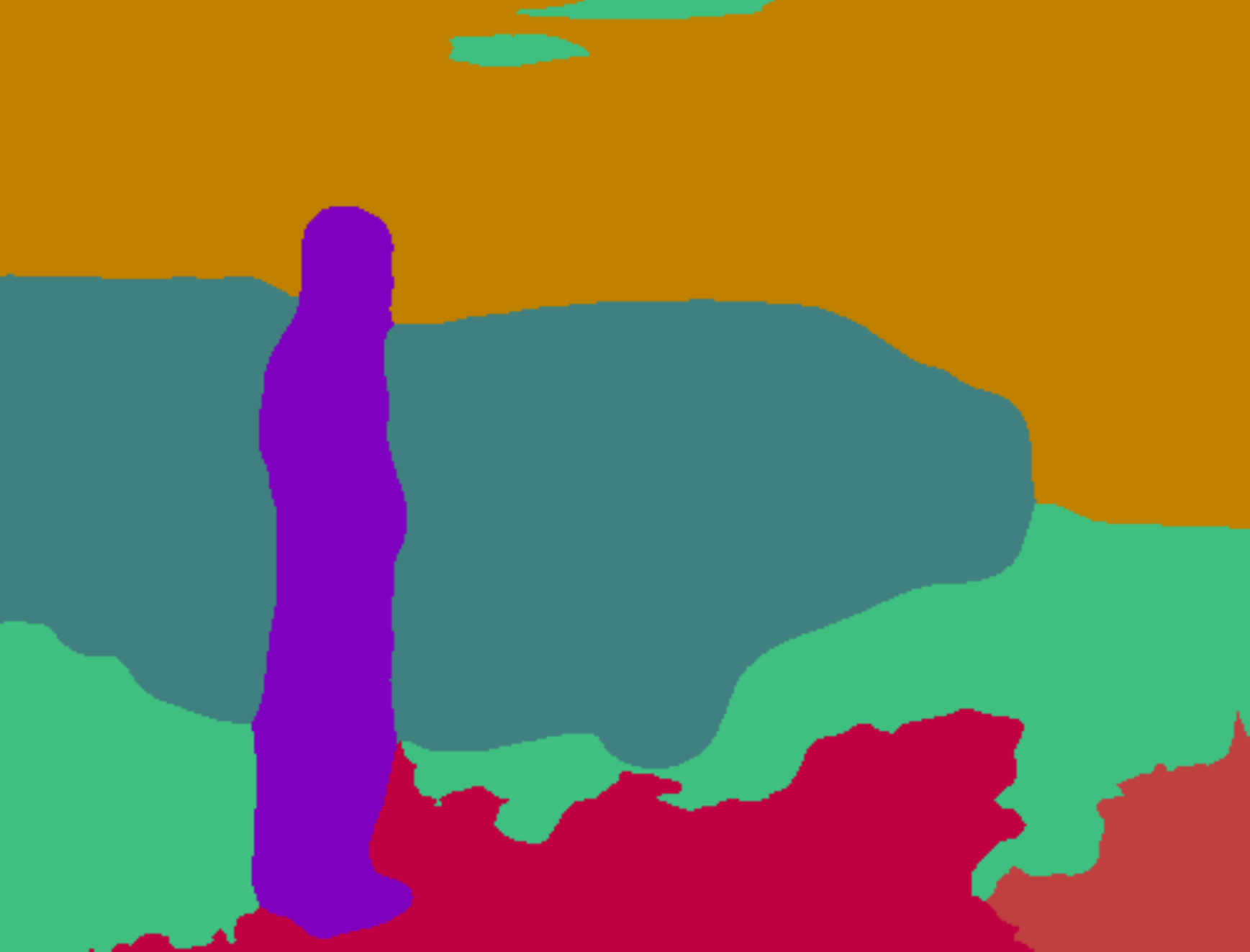} &
    \includegraphics[width=2.50cm]{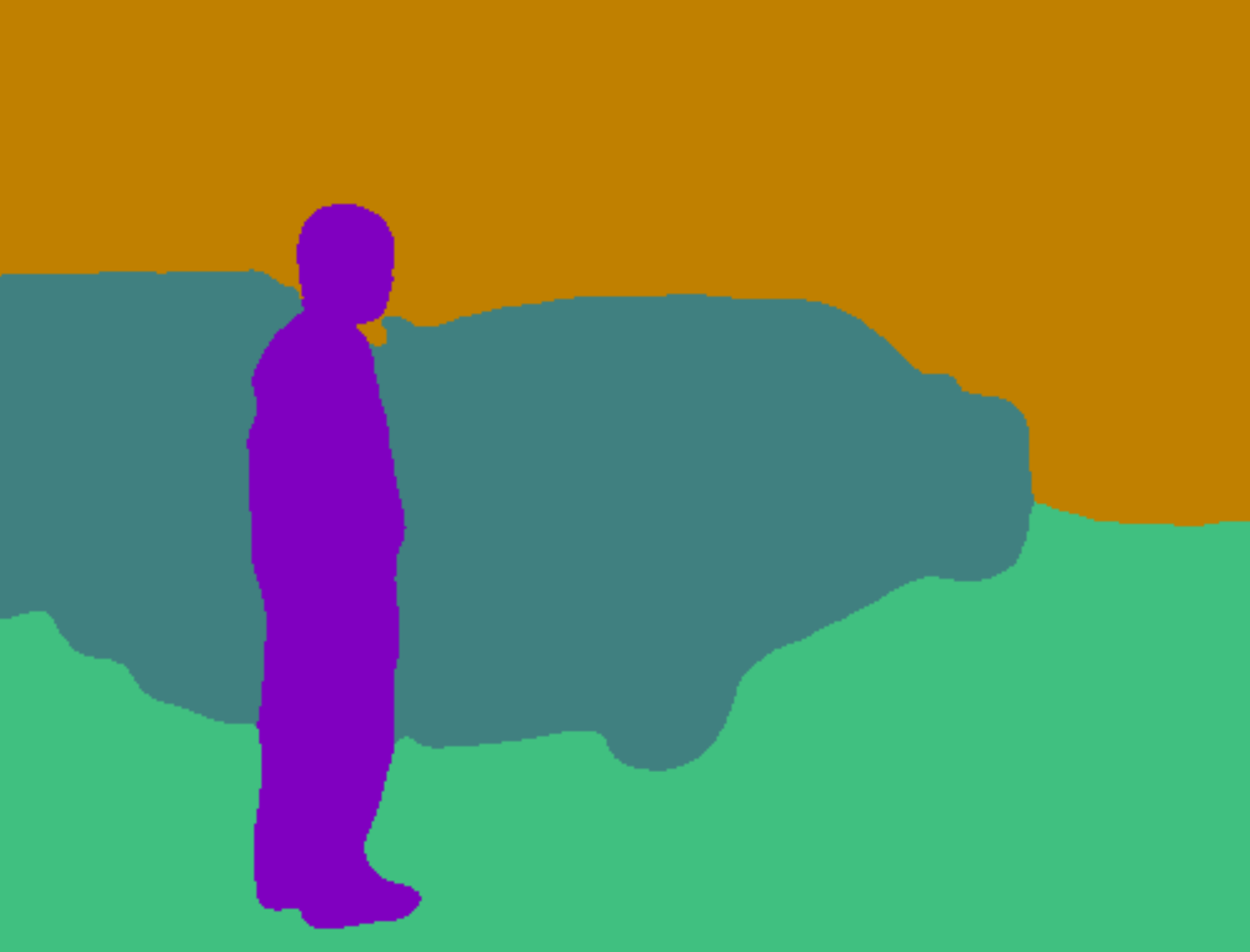} \\
    \includegraphics[width=2.50cm]{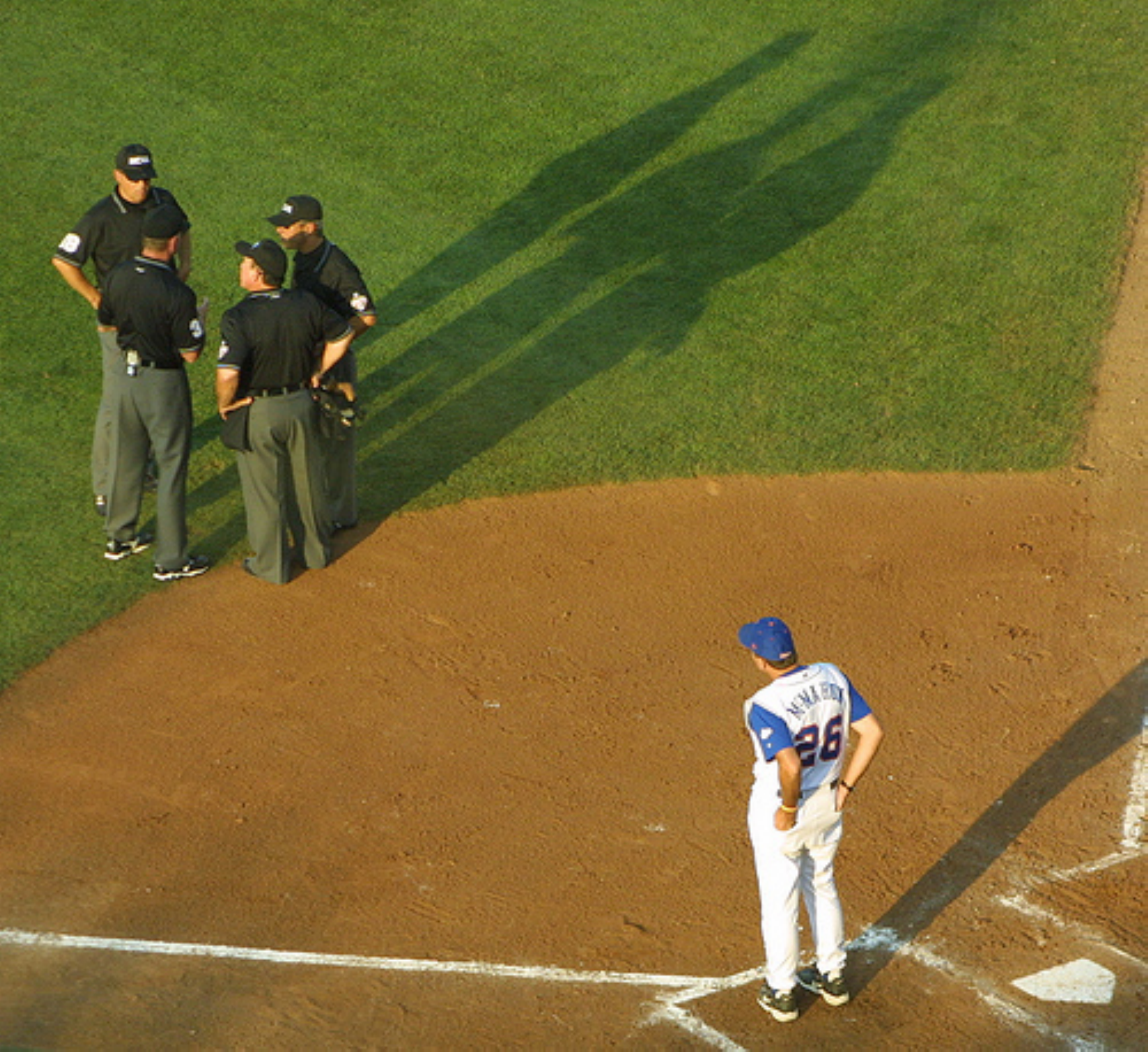} &
    \includegraphics[width=2.50cm]{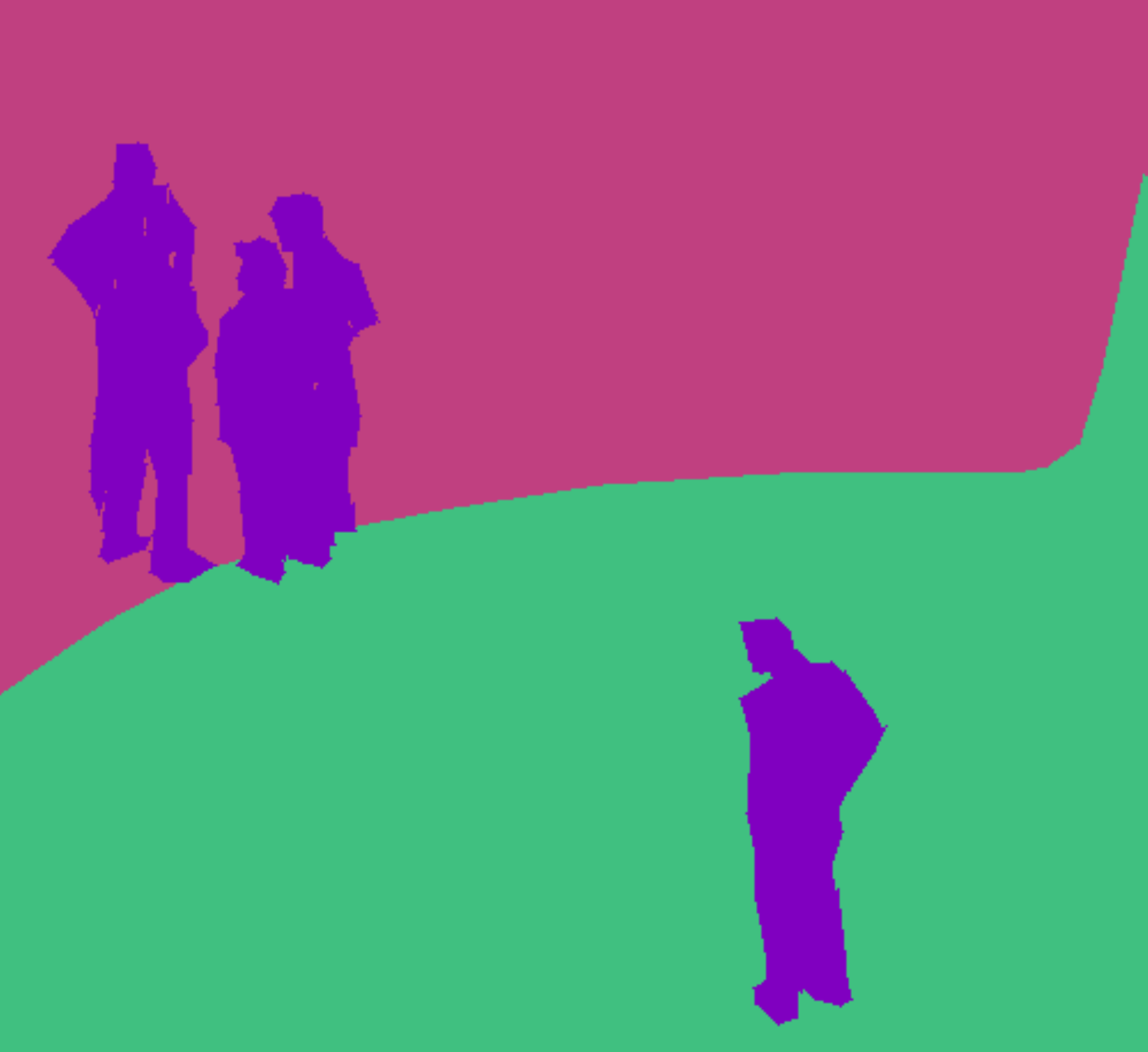} &
    \includegraphics[width=2.50cm]{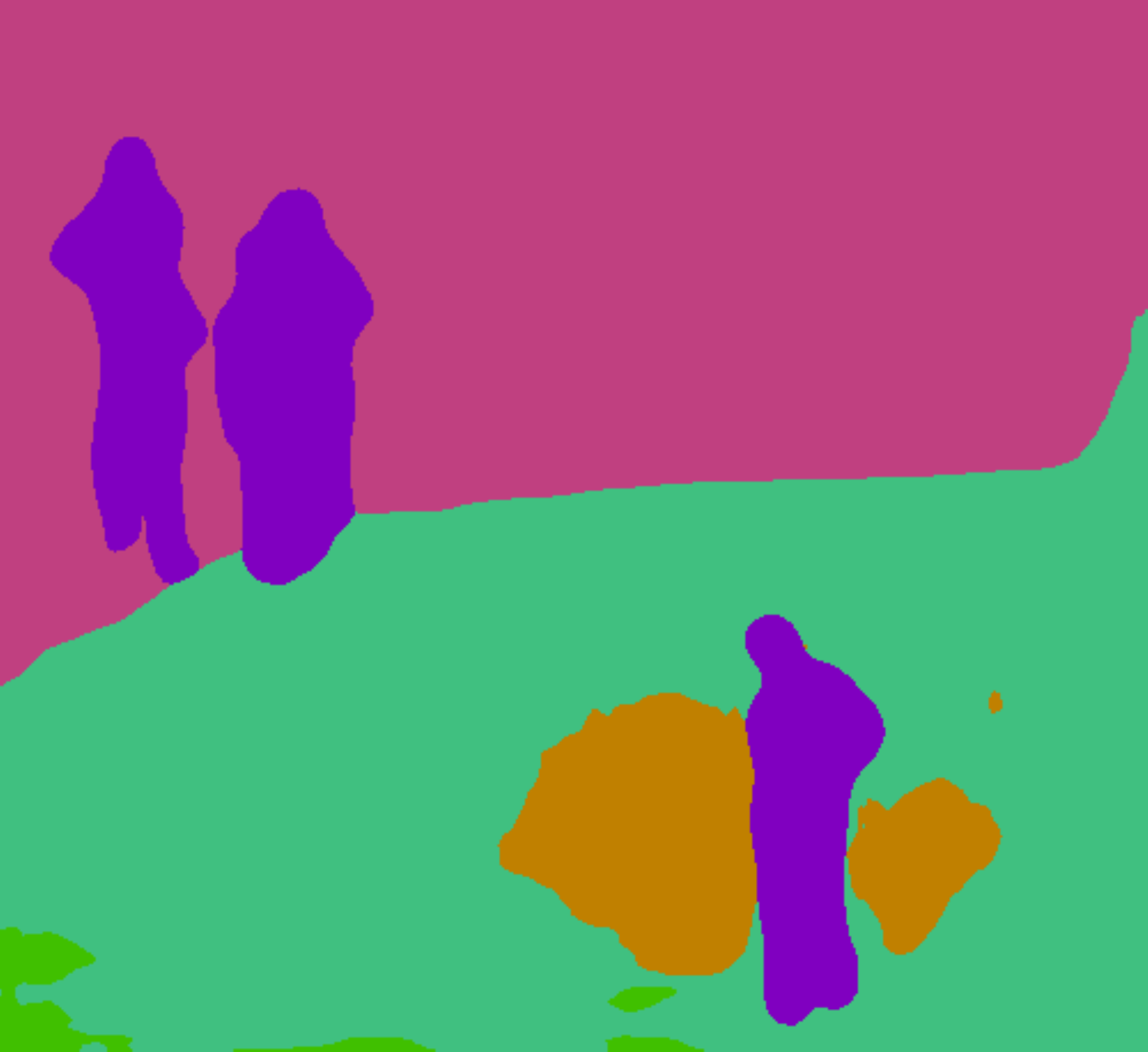} &
    \includegraphics[width=2.50cm]{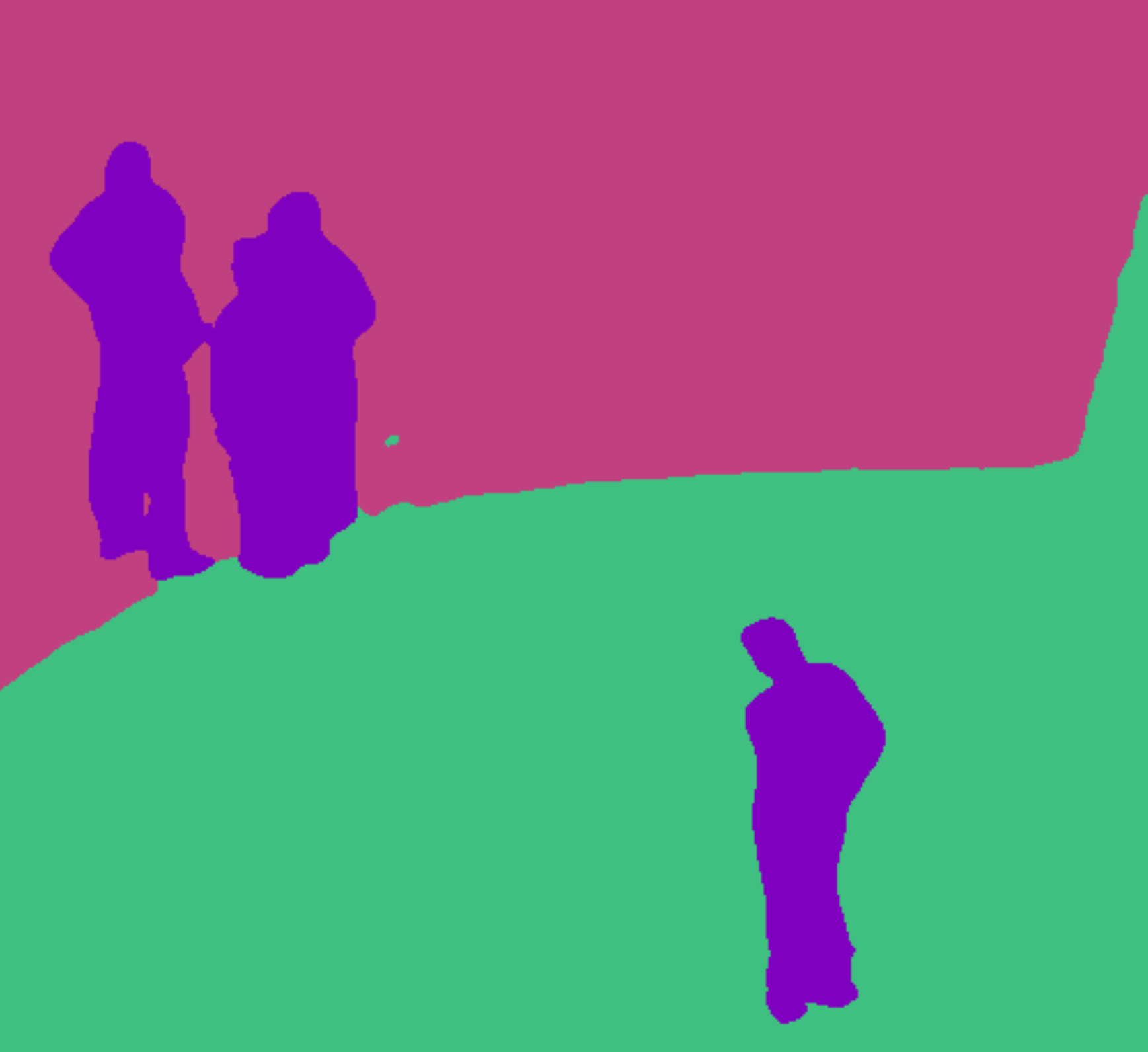} \\
    \includegraphics[width=2.50cm]{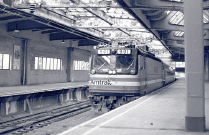} &
    \includegraphics[width=2.50cm]{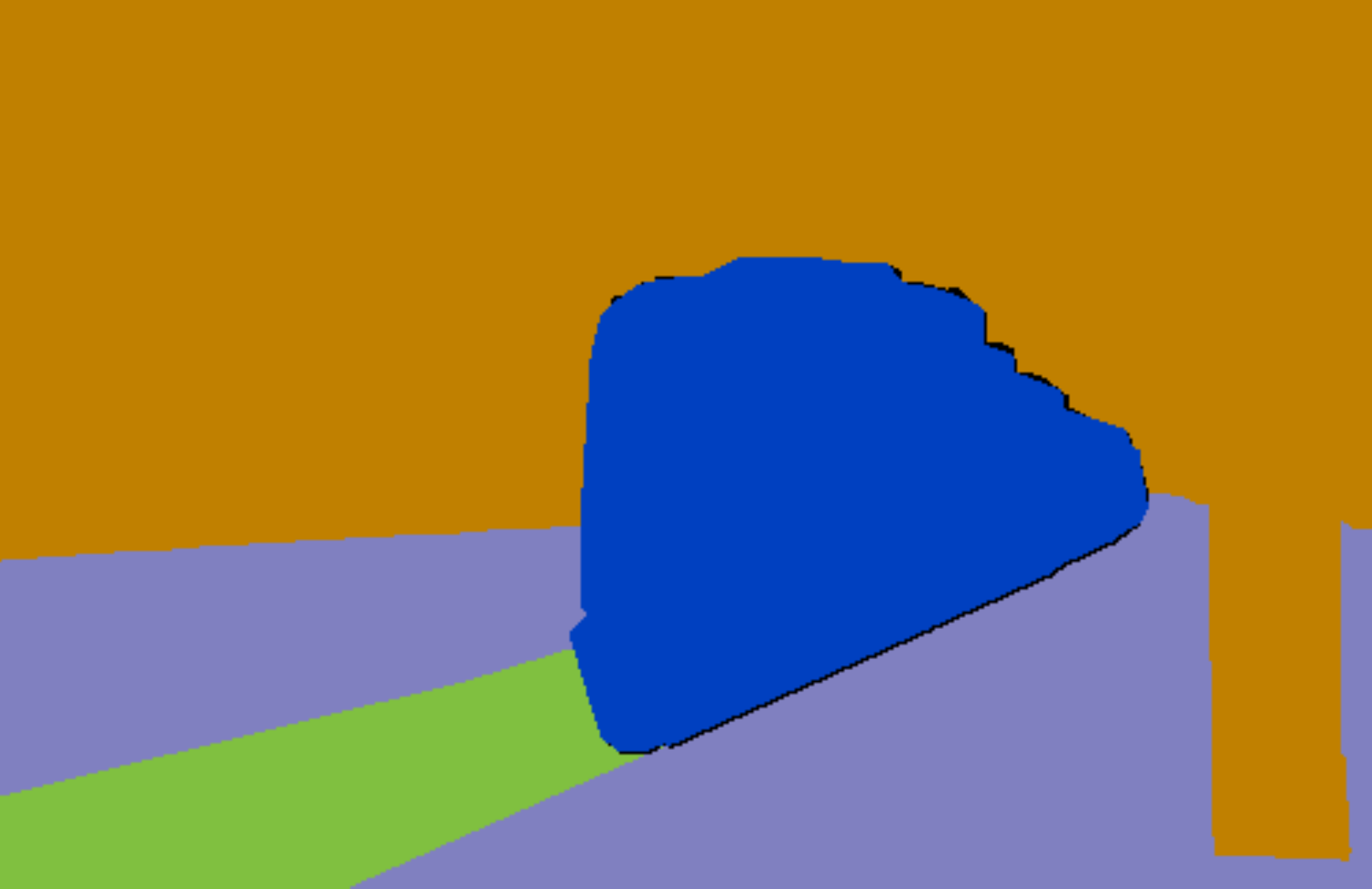} &
    \includegraphics[width=2.50cm]{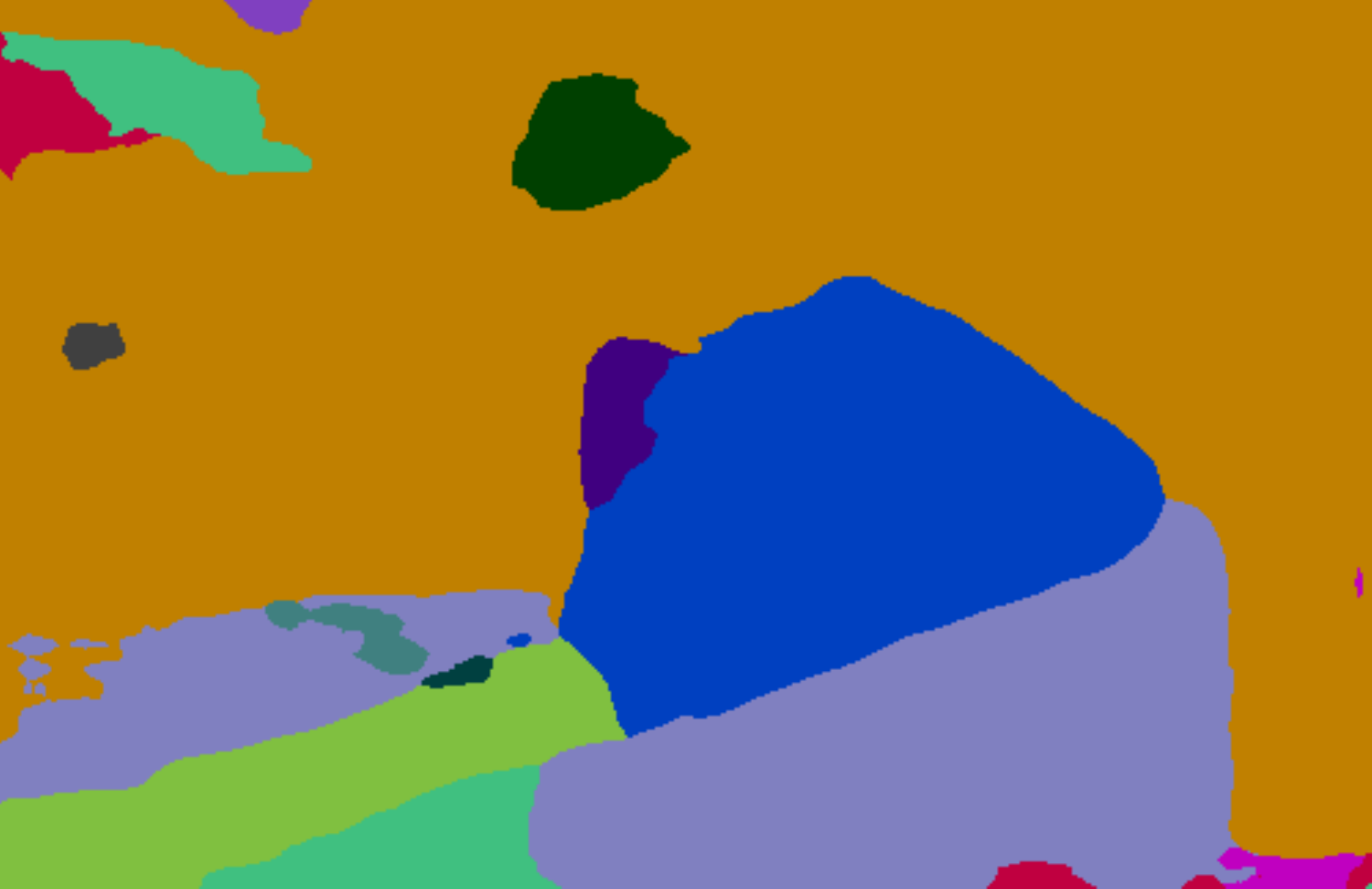} &
    \includegraphics[width=2.50cm]{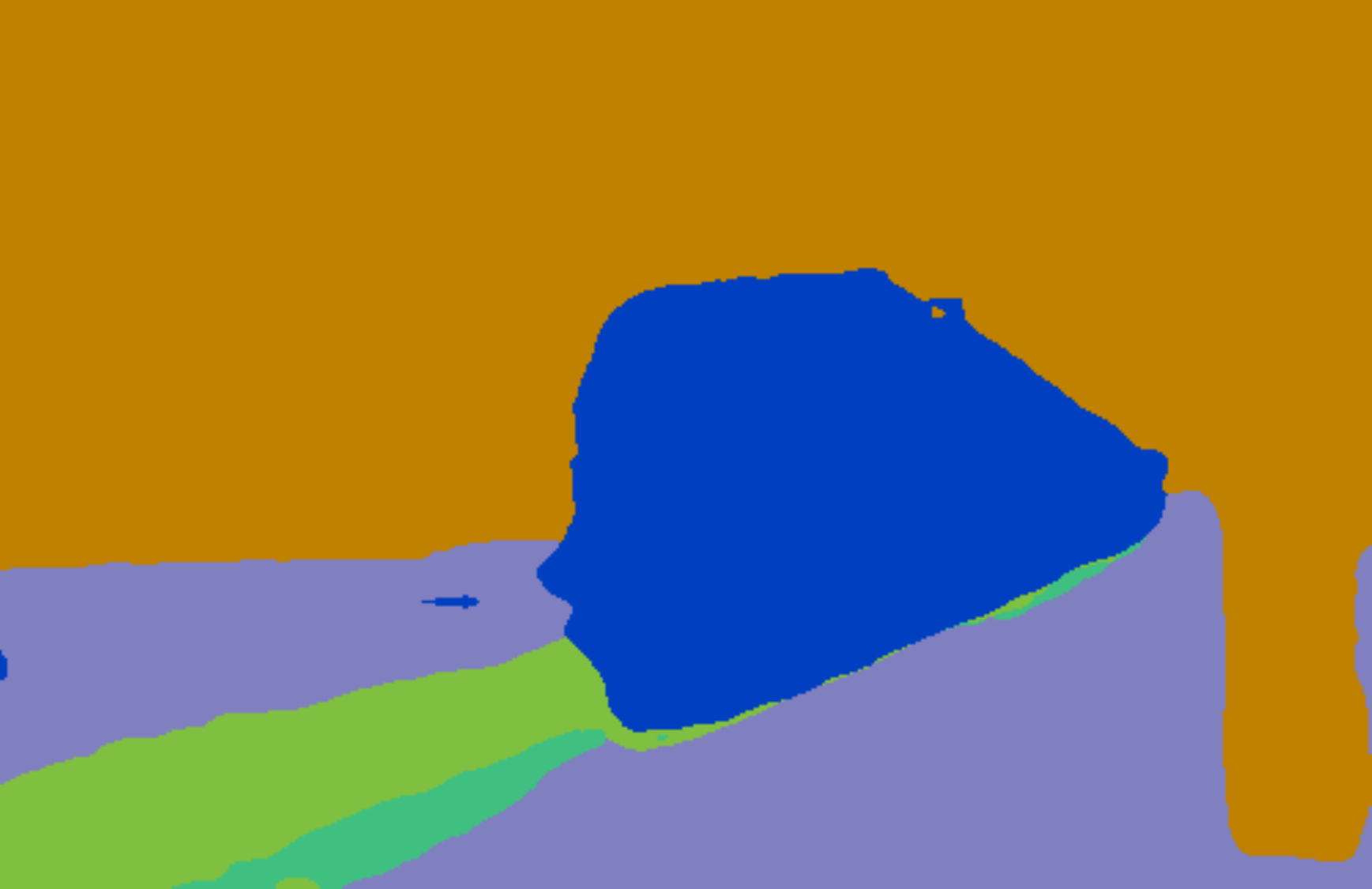} \\
    \centering (a) Image & (b) GT & (c) Baseline & (d) EfficientFCN \\
\end{tabular}
\end{center}
\caption{Visualization results from the PASCAL Context dataset.}
\label{fig:vis}
\end{figure}
%\end{comment}

%\vspace{4pt}
\noindent \textbf{Number of holistic codewords.} We also conduct  experiments to survey the
effectiveness of the number of codewords in our predicted semantic codebook for feature upsampling.
As shown in Table \ref{table:ablation_n_codewords}, as the number of the semantic codewords
increases from 32 to 512, the performance improves 1\% in terms of mIoU on PASCAL Context.
However, when the number of the semantic codewords further increases from 512 to 1024, the performance
has a slight drop, which might be caused by the additional parameters. The larger model capacity
might cause model to overfit the training data. In addition, since the assembly coefficients of the
semantic codewords are predicted from the OS=8 multi-scale fused feature $m_8$, the increased number
of the semantic codewords also leads to significantly more extra computational cost. Thus, to balance
the performance and also the efficiency, we set the number of the holistic codewords as 256 for the
PASCAL Context and PASCAL VOC 2012 datasets. Since PASCAL Context only has 60 classes and we observe
the number of codewords needed is approximately 4 times than the number of classes. We therefore set
the number of codewords as 600 for ADE20K, which has 150 classes.

%\vspace{4pt}
\noindent \textbf{Importance of the codeword information transfer for accurate assembly coefficient estimation.} 
The key of our proposed HGD is how to linearly assemble holistic codewords at each spatial location
to form high-resolution upsampled feature maps based on the feature maps $m_8$. In our HGD, although
the OS=8 features have well maintained structural image information, we argue that directly using OS=8 features to predict codeword assembly coefficients are less effective since they have no information about the codewords. 
We propose to transfer the codeword information as the average codeword basis,
which is location-wisely added to the OS=8 feature maps. To verify this
argument, we design an experiment that removes the additive information
transfer, and only utilizes two $1\times 1$ convolutions with the same output
channels on the OS=8 feature maps $m_8$ for directly predicting assembly
coefficients. The mIoU of this implementation is 54.2\%, which has a clear performance drop if there
is no codeword information transfer from the codeword generation branch to the codeword coefficient prediction branch.
%that reconstructing high-resolution feature representations from holistic codewords that capture various contexts by holistic codewords assembly. Thus, it is critical to generate coefficients of codewords assembly for each position of the reconstructed high-resolution feature representations. it is not enough and reasonable to predict the coefficients of codewords assembly by simply applying a convolution (1*1) to the fused high-resolution feature representations from different stages of the backbone network (OS=8, OS=16, and OS=32). Although the fused high-resolution feature representations contain rich low-level detail information and high-level semantic information, it is still lacking the knowledge of holistic codewords which is import for predicting the coefficients of codewords assembly. The results as shown in Table XX demonstrate the importance of adding the knowledge of holistic codewords to the fused high-resolution feature representations to predict coefficients for high-resolution feature representations by codewords assembly. (more detailed results)
%\noindent {\bf Ablation study on HED.} 
%blablabla....
% \noindent {\bf Ablation study with the Encoder-Decoder.} 
% blablabla....
% \noindent {\bf Comparisons with the state-of-art methods.} 
% blablabla....

%\vspace{4pt}
\noindent \textbf{Visualization of the weighting maps and example results.} 
%To better interpret the obtained holistic codewords, we visualize the assembly coefficient maps, which are used for holistic codewords generation. As shown in Figure \ref{fig:weighting_maps}, the
%different regions of the weighting maps are activated for sensing 
%class-agnostic regions. With such weighting maps, the globally weighting with the semanitc codebook base, the different codewords are created from its beloning and can capture the calss-agnostic semantic information. NEED FINAL RESULTS TO WRITE THIS PARAGRAPH.
To better interpret the obtained holistic codewords, we visualize the weighting maps $\tilde{A}$ for
creating the holistic codewords in Fig.~\ref{fig:weighting_maps}, where each column shows one
weighting map $\tilde{A}_i$ for generating one holistic codeword. Some weighting maps focus on summarizing  foreground objects or regions to create holistic codewords, while some other weighting maps pay attention to summarizing background contextual regions or objects as the holistic codewords. The visualization shows that the learned codewords implicitly capture different global contexts from the scenes.
In Fig.~\ref{fig:vis}, we also visualize some predictions by the baseline
DilatedFCN-8s and by our EfficientFCN, where our model significantly improves the visualized results with the proposed HGD.

\begin{table*}[!t]
\small
\centering
\caption{Results of each category on PASCAL VOC 2012 test set. Our
        EfficientFCN obtains 85.4 \% without MS COCO dataset pre-training and 87.6\% with MS COCO dataset pre-training. (For each
    columns, the best two entries are filled in gray color. )}
\label{table:pascal_voc_2012}
\resizebox{\textwidth}{!}{%
\begin{tabular}{l|cccccccccccccccccccc|c}
\hline
\textbf{Method}    & \textbf{aero} & \textbf{bike} & \textbf{bird} & \textbf{boat} & \textbf{bottle} & \textbf{bus}  & \textbf{car}  & \textbf{cat}  & \textbf{chair} & \textbf{cow}  & \textbf{table} & \textbf{dog}  & \textbf{horse} & \textbf{mbike} & \textbf{person} & \textbf{plant} & \textbf{sheep} & \textbf{sofa} & \textbf{train} & \textbf{tv}   & \textbf{mIoU\%} \\ \hline\hline
\textbf{FCN} \cite{long2015fully}       & 76.8          & 34.2          & 68.9
& 49.4          & 60.3            & 75.3          & 74.7          & 77.6
& 21.4           & 62.5          & 46.8           & 71.8          & 63.9
& 76.5           & 73.9            & 45.2           & 72.4           & 37.4
& 70.9           & 55.1          & 62.2   \\
\textbf{DeepLabv2} \cite{chen2017deeplab} & 84.4          & 54.5          & 81.5
& 63.6          & 65.9            & 85.1          & 79.1          & 83.4
& 30.7           & 74.1          & 59.8           & 79.0            & 76.1           & 83.2           & 80.8            & 59.7           & 82.2           & 50.4          & 73.1           & 63.7          & 71.6  \\
\textbf{CRF-RNN} \cite{CRF-RNN}   & 87.5          & 39.0          & 79.7          & 64.2          & 68.3            & 87.6          & 80.8          & 84.4          & 30.4           & 78.2          & 60.4           & 80.5          & 77.8           & 83.1           & 80.6            & 59.5           & 82.8           & 47.8          & 78.3           & 67.1          & 72.0  \\
\textbf{DeconvNet} \cite{DeconvNet} & 89.9          & 39.3          & 79.7          & 63.9          & 68.2            & 87.4          & 81.2          & 86.1          & 28.5           & 77.0          & 62.0           & 79.0          & 80.3           & 83.6           & 80.2            & 58.8           & 83.4           & 54.3          & 80.7           & 65.0            & 72.5   \\
\textbf{DPN} \cite{DPN}       & 87.7          & 59.4          & 78.4          & 64.9          & 70.3            & 89.3          & 83.5          & 86.1          & 31.7           & 79.9          & 62.6           & 81.9          & 80.0           & 83.5           & 82.3            & 60.5           & 83.2           & 53.4          & 77.9           & 65.0            & 74.1   \\
\textbf{Piecewise} \cite{Piecewise} & 90.6          & 37.6          & 80.0
& 67.8          & 74.4            & 92            & 85.2          & 86.2
& 39.1           & 81.2          & 58.9           & 83.8          & 83.9
& 84.3           & 84.8            & 62.1           & 83.2           & 58.2
& 80.8           & 72.3          & 75.3    \\
\textbf{ResNet38} \cite{ResNet38}  & 94.4          & 72.9          & 94.9
& 68.8          & 78.4            & 90.6          & 90.0          & 92.1
& 40.1           & 90.4          & 71.7           & 89.9          & 93.7
& \bgGray 91.0           & 89.1            & 71.3           & 90.7           & 61.3          & 87.7           & 78.1          & 82.5     \\
\textbf{PSPNet} \cite{zhao2017pyramid}    & 91.8          & 71.9          & 94.7          & 71.2          & 75.8            & 95.2          & 89.9          & 95.9          & 39.3           & 90.7          & 71.7           & 90.5          & 94.5           & 88.8           & 89.6            & 72.8           & 89.6           & \bgGray {64.0}          & 85.1           & 76.3          & 82.6   \\
\textbf{EncNet} \cite{Zhang_2018_CVPR}    & 94.1          & 69.2          & \bgGray\textbf{96.3} & \bgGray 76.7          & \bgGray \textbf{86.2}   & 96.3          & 90.7          & 94.2          & 38.8           & 90.7          & 73.3           & 90.0          & 92.5           & 88.8           & 87.9            & 68.7           & 92.6           & 59.0          & 86.4           & 73.4          & 82.9            
            \\
            \textbf{APCNet} \cite{he2019adaptive}      & 95.8 &\bgGray 75.8 & 84.5  & 76.0 & 80.6 &
            \bgGray 96.9 & 90.0 & 96.0 & \bgGray\textbf{42.0} & \bgGray 93.7
            &\bgGray 75.4 & 91.6 & 95.0 & 90.5 &
             89.3 &  75.8 & 92.8 & 61.9 &  88.9 & \bgGray 79.6 & 84.2
            \\
            \textbf{CFNet} \cite{Zhang_2019_CVPR}      & 95.7 & 71.9 &\bgGray
            95.0  &\bgGray 76.3 & \bgGray 82.8 &
            94.8 & 90.0 & 95.9 & 37.1 & 92.6 & 73.0 & \bgGray 93.4 & 94.6 & 89.6 &
            88.4 & 74.9 & \bgGray \textbf{95.2} &  63.2 & \bgGray \textbf{89.7} & 78.2 & 84.2
            \\
            \textbf{DMNet} \cite{he2019dynamic}        & \bgGray 96.1 &
            \bgGray\textbf{77.3} & 94.1 & 72.8 & 78.1 & 
            \bgGray\textbf{97.1} & \bgGray \textbf{92.7} & \bgGray 96.4 & 39.8 & 91.4 & \bgGray 75.5 & 92.7 & \textbf{95.8} &
            \bgGray {91.0} & \bgGray {90.3} & \bgGray {76.6} & \bgGray 94.1 & 62.1 & 85.5 & 77.6 & \bgGray 84.4 
            \\ \hline
%\textbf{PAN}       & 95.7          & 75.2          & 94.0          & 73.8          & 79.6            & 96.5          & \textbf{93.7} & 94.1          & 40.5           & 93.3          & 72.4           & 89.1          & 94.1           & \textbf{91.6}  & \textbf{89.5}   & 73.6           & \textbf{93.2}  & \textbf{62.8} & 87.3           & 78.6          & 84.0            \\ \hline
            \textbf{Ours}      & \bgGray \textbf{96.4} & {74.1} &
            92.8 & \bgGray 75.6 & 81.9 &\bgGray 96.9 
            & \bgGray {92.6}  & \bgGray \textbf{97.1} & \bgGray 41.6 & \bgGray \textbf{95.4} 
            & 72.9 & \bgGray \textbf{93.9} & \bgGray \textbf{95.9} 
            & {90.6} &\bgGray \textbf{ 90.6} &\bgGray \textbf{77.2}  & 94.0 &
            67.5 &\bgGray 89.3 &
            \bgGray \textbf{79.8} & \bgGray \textbf{85.4} \\
             \hline
            \multicolumn{22}{c}{\textbf{With COCO Pre-training}}\\
            \hline
            
            \textbf{CRF-RNN}~\cite{CRF-RNN} & 90.4 & 55.3 & 88.7 & 68.4 & 69.8 & 88.3 & 82.4 & 85.1 & 32.6 & 78.5 & 64.4 & 79.6 & 81.9 & 86.4 & 81.8 & 58.6 & 82.4 & 53.5 & 77.4 & 70.1 & 74.7 \\
            %BoxSup~\cite{dai2015boxsup} & 89.8 & 38.0 & 89.2 & 68.9 & 68.0 & 89.6 & 83.0 & 87.7 & 34.4 & 83.6 & 67.1 & 81.5 & 83.7 & 85.2 & 83.5 & 58.6 & 84.9 & 55.8 & 81.2 & 70.7 & 75.2 \\
            \textbf{Piecewise}~\cite{Piecewise} & 94.1 & 40.7 & 84.1 & 67.8 & 75.9 & 93.4 & 84.3 & 88.4 & 42.5 & 86.4 & 64.7 & 85.4 & 89.0 & 85.8 & 86.0 & 67.5 & 90.2 & 63.8 & 80.9 & 73.0 & 78.0 \\
            %FCRNs~\cite{wu2016bridging} & 91.9 & 48.1 & 93.4 & 69.3 & 75.5 & 94.2 & 87.5 & 92.8 & 36.7 & 86.9 & 65.2 & 89.1 & 90.2 & 86.5 & 87.2 & 64.6 & 90.1 & 59.7 & 85.5 & 72.7 & 79.1 \\
            %LRR~\cite{ghiasi2016laplacian} & 92.4 & 45.1 & 94.6 & 65.2 & 75.8 & {95.1} & 89.1 & 92.3 & 39.0 & 85.7 & 70.4 & 88.6 & 89.4 & 88.6 & 86.6 & 65.8 & 86.2 & 57.4 & 85.7 & 77.3 & 79.3 \\
            \textbf{DeepLabv2}~\cite{chen2017deeplab} & 92.6 & 60.4 & 91.6 & 63.4 & 76.3 & 95.0 & 88.4 & 92.6 & 32.7 & 88.5 & 67.6 & 89.6 & 92.1 & 87.0 & 87.4 & 63.3 & 88.3 & 60.0 & 86.8 & 74.5 & 79.7  \\
            \textbf{RefineNet}\cite{RefineNet}  & 95.0 & 73.2 & 93.5 & 78.1 & 84.8 & 95.6 & 89.8 & 94.1 & 43.7 & 92.0 & 77.2 & 90.8 & 93.4 & 88.6 & 88.1 & 70.1 & 92.9 & 64.3 & 87.7 & 78.8 & 84.2 \\
            \textbf{ResNet38}\cite{ResNet38}  &  96.2 & 75.2 & \secbest
            95.4 & 74.4 & 81.7 & 93.7 & 89.9 & 92.5 & \secbest 48.2 & 92.0 & 79.9
            & 90.1 & 95.5 & 91.8 & 91.2 &  73.0 & 90.5 & 65.4 & 88.7 & 80.6 & 84.9\\
            \textbf{PSPNet}~\cite{zhao2017pyramid} & {95.8} & {72.7} & {95.0}
            & {78.9} & {84.4} & 94.7 & \secbest{92.0} & {95.7} & {43.1} &
            {91.0} & \best{80.3} & {91.3} & {96.3} & {92.3} & {90.1} & {71.5}
            & \secbest{94.4} & \secbest{66.9} & {88.8} & \best{82.0} & {85.4} \\
            \textbf{DeepLabv3}\cite{chen2017rethinking} & \secbest 96.4 & 76.6
            & 92.7 & 77.8 & \secbest{87.6} & 96.7 & 90.2 & 95.4 &  47.5 &
            \secbest 93.4 & 76.3 & 91.4 & \best{97.2} &  91.0 & \best{92.1} &
            71.3 & 90.9 & \secbest{68.9} & \secbest{90.8} & 79.3 & 85.7 \\ 
            \textbf{EncNet}\cite{Zhang_2018_CVPR}  & 95.3 & 76.9 & 94.2 &
            \secbest 80.2 & 85.2 & 96.5 & 90.8 & 96.3 &  47.9
            &  93.9 & \secbest 80.0 & 92.4 & \secbest 96.6 & 90.5 &  91.5 & 70.8 &  93.6 & 66.5 & 87.7 & 80.8 & 85.9 
            %\hline
            \\ 
            \textbf{CFNet} \cite{Zhang_2019_CVPR}      &\best 96.7 &\secbest 79.7 &
            94.3  & 78.4 & 83.0 & \best 97.7 & 91.6 &\secbest 96.7 &\best 50.1
            &\secbest 95.3 & 79.6 & \bgGray 93.6 &\best 97.2 &\secbest
            94.2 &\secbest 91.7 & \bgGray {78.4} &\best  95.4 & \bgGray
            \textbf{69.6} & 90.0 & 81.4 & \secbest 87.2
            \\ \hline
            \textbf{Ours} & \bgGray {96.6} & \best {80.6} &
            \best 96.1 & \best{82.3} &\best 87.8 &\best 97.7 
            & \best {94.4}  & \best {97.3} &  47.1 & \bgGray \textbf{96.3} 
            & {77.9} & \bgGray \textbf{94.8} & \bgGray
            \textbf{97.2} 
            &\bgGray \textbf{94.3} & 91.1 & \best 81.0  & 94.3 & 61.5
            &\best 91.6 &\best {83.5} & \bgGray \textbf{87.6}
            \\ \hline
\end{tabular}}
\end{table*}

%\vspace{6pt}
\noindent
\textbf{Comparison with state-of-the-art methods.} 
To further demonstrate the effectiveness of our proposed EffectiveFCN with the holistically-guided decoder, the comparisons with state-of-the-art methods are shown in Table \ref{table:pascal-context}. The dilatedFCN based methods dominate semantic segmentation. However, our work is still able to achieve the best results compared to the dilatedFCN based methods on the PASCAL Context validation set without using any dilated convolution and has significantly less computational cost. Because of the efficient design of our HGD, our EfficientFCN only has 1/3 of the computational cost of state-of-the-arts methods but can still achieve the best performance. 
%\hfill

\subsection{Results on PASCAL VOC}
The original PASCAL VOC 2012 dataset consists of 1,464 images for training, 1,449 for validation, and 1,456 for testing, which is a major benchmark dataset for semantic object segmentation. It includes 20 foreground objects classed and one background class. The augmented training set of 10,582 images, namely train-aug, is adopted as the training set following the previous experimental set in \cite{Zhang_2019_CVPR}.
To further demonstrate the effectiveness of our proposed HGD. We adopt all the best strategies of HGD design and compare it with state-of-the-art methods on the test set of PASCAL-VOC 2012, which is evaluated on the official online server. As shown in Table \ref{table:pascal_voc_2012}, the dilatedFCN based methods dominate the top performances on the PSCAL VOC benchmark. However, our EfficientFCN with a backbone having no dilated convolution can still achieve the best results among all the ResNet101-based methods. 
\subsection{Results on ADE20K}
The ADE20K dataset consists of 20K images for training, 2K images for
validation, and 3K images for testing, which were used for ImageNet Scene
Parsing Challenge 2016. This dataset is more complex and challenging with 150 labeled classes and more diverse scenes. As shown in Table \ref{table:pascal-context}, our
EfficientFCN achieves the competitive performance than the dilatedFCN based
methods but has only 1/3 of their computational cost.
\section{Conclusions}
In this paper, we propose the EfficientFCN model with the holistically-guied decoder for achieving efficient and accurate semantic segmentation. The novel decoder is able to reconstruct the high-resolution semantic-rich feature maps from multi-scale feature maps of the encoder. 
%With the proposed HGD, we implement a novel method efficientFCN for the semantic segmentation. 
Because of the superior feature upsampling performance of the HGD, our EfficientFCN, with much fewer parameters and less computational cost, achieves competitive or even better performance compared with state-of-the-art dilatedFCN based methods.

\subsection*{Acknowledgements}
This work is supported in part by SenseTime Group Limited, in part by the General Research Fund through the Research Grants Council of Hong Kong under Grants CUHK 14202217 / 14203118 / 14205615 / 14207814 / 14213616 / 14208417 / 14239816, in part by CUHK Direct Grant.

\clearpage

%%%%%%%%%%%%%%%%%%%%%%%

{\small
%\bibliographystyle{ieee_fullname}
%\bibliography{semanticseg_cvpr20}
\bibliographystyle{splncs04}
\bibliography{5211}
}

\end{document}